\documentclass[lettersize,journal]{IEEEtran}  % Comment this line out if you need a4paper
\usepackage{amsmath,amsfonts}
\usepackage{array}
\usepackage{textcomp}
\usepackage{stfloats}
\usepackage{url}
\usepackage{verbatim}
\usepackage{graphicx}
\usepackage{cite}
\usepackage{bm}
\usepackage{amsmath}
\usepackage{cite}
\usepackage{amsmath,amssymb,amsfonts}
\usepackage{ifthen}
\usepackage[numbers]{natbib}
\usepackage{multicol}
\usepackage[bookmarks=true]{hyperref}
\usepackage{graphicx}
\usepackage{textcomp}
\usepackage{stfloats}
\usepackage{graphicx}
\usepackage{booktabs}
\usepackage{float}
\usepackage{listings}
\usepackage{url}
\usepackage{algpseudocode}
\usepackage[ruled,vlined]{algorithm2e}
\usepackage{multirow}
\usepackage{mathrsfs}
\usepackage{amsmath,bm}
\usepackage[switch]{lineno}
\usepackage[table,xcdraw]{xcolor}
\usepackage{comment}
\usepackage{hyperref}
\usepackage{subfigure}
\usepackage{stfloats}
\usepackage{circuitikz}
\usepackage{tikz}
\usepackage{ifthen}
\usepackage{bm}
\usepackage{makecell}
\usepackage{multirow}

\begin{document}

%%%%% Title
\title{Real-Time Metric-Semantic Mapping for Autonomous Navigation in Outdoor Environments}
\author{
	Jianhao Jiao,
	Ruoyu Geng,
	Yuanhang Li,
	Ren Xin,
	Bowen Yang,
	Jin Wu,\\
	Lujia Wang,
	Ming Liu,
	Rui Fan, %~\IEEEmembership{Senior Member,~IEEE},
	Dimitrios Kanoulas
	\thanks{This work was supported by the National Natural Science Foundation of China (Grants No. 62303388 and 62233013), UK Research and Innovation Future Leaders Fellowship (RoboHike, Grant No. MR/V025333/1), Shanghai Municipal Science and Technology Major Project (Grant No. 2021SHZDZX0100), and Xiaomi Young Talents Program. (\textit{Corresponding author: Jianhao Jiao.})}
	\thanks{Jianhao Jiao and Dimitrios Kanoulas are with the Department of Computer Science, University College London, WC1E 6BT London, U.K. (e-mail: ucacjji@ucl.ac.uk; d.kanoulas@ucl.ac.uk).}
	\thanks{Ruoyu Geng, Ren Xin, and Ming Liu are with the Robotics and Autonomous Systems, The Hong Kong University of Science and Technology (Guangzhou), Nansha, Guangzhou, Guangdong 511400, China.}
        \thanks{Yuanhang Li, Bowen Yang, and Jin Wu are with the Department of Electronic and Computer Engineering, The Hong Kong University of Science and Technology, Hong Kong, SAR, China.}
        \thanks{Lujia Wang is with The Hong Kong University of Science and Technology (Guangzhou), Nansha, Guangzhou, Guangdong 511400, China.}
	\thanks{Rui Fan is with the College of Electronics and Information Engineering, Shanghai Research Institute for Intelligent Autonomous Systems, the State Key Laboratory of Intelligent Autonomous Systems, and the Frontiers Science Center for Intelligent Autonomous Systems, Tongji University, Shanghai 201804, China (e-mail: rui.fan@ieee.org).}
}

\maketitle

%%%%% Content
\begin{abstract}  
    The creation of a metric-semantic map, which encodes human-prior knowledge, represents a high-level abstraction of environments.
    However, constructing such a map poses challenges related to the fusion of multi-modal sensor data, the attainment of real-time mapping performance, and the preservation of structural and semantic information consistency.
    In this paper, we introduce an online metric-semantic mapping system that utilizes LiDAR-Visual-Inertial sensing to generate a global metric-semantic mesh map of large-scale outdoor environments.
    Leveraging GPU acceleration, our mapping process achieves exceptional speed, with frame processing taking less than $7ms$, regardless of scenario scale.
    Furthermore, we seamlessly integrate the resultant map into a real-world navigation system, enabling metric-semantic-based terrain assessment and autonomous point-to-point navigation within a campus environment. Through extensive experiments conducted on both publicly available and self-collected datasets comprising $24$ sequences, we demonstrate the effectiveness of our mapping and navigation methodologies.
    Code has been publicly released: \url{https://github.com/gogojjh/cobra}.
\end{abstract}
\vspace{-0.1cm}

% 1. Is the NtP addressed to industrial practitioners and sufficiently distinct from the Abstract (which is addressed to colleagues in research)?
% 2. Does the NtP summarize the innovative aspects of the paper, the practical value of the results, and how the results might be applied in the near or mid-term from a practitioner's point of view?
% 3. Does the NtP clearly describe practical limitations of the approach and directions for future research?
\def\abstractname{Note to Practitioners}
\begin{abstract}
    This paper tackles the challenge of autonomous navigation for mobile robots in complex, unstructured environments with rich semantic elements. Traditional navigation relies on geometric analysis and manual annotations, struggling to differentiate similar structures like roads and sidewalks. We propose an online mapping system that creates a global metric-semantic mesh map for large-scale outdoor environments, utilizing GPU acceleration for speed and overcoming the limitations of existing real-time semantic mapping methods, which are generally confined to indoor settings. Our map integrates into a real-world navigation system, proven effective in localization and terrain assessment through experiments with both public and proprietary datasets. Future work will focus on integrating kernel-based methods to improve the map's semantic accuracy.
\end{abstract}
\vspace{-0.1cm}


\vspace{-0.1cm}
\section{Introduction}

%%%%%%%%%%%%%%%%%%%%%%%%%%%%%%%%%%%%%%%%%%%%%%%%%%%%%%%%%%%%%%%%%
\begin{figure}[t]
      \centering
      \includegraphics[width=0.45\textwidth]{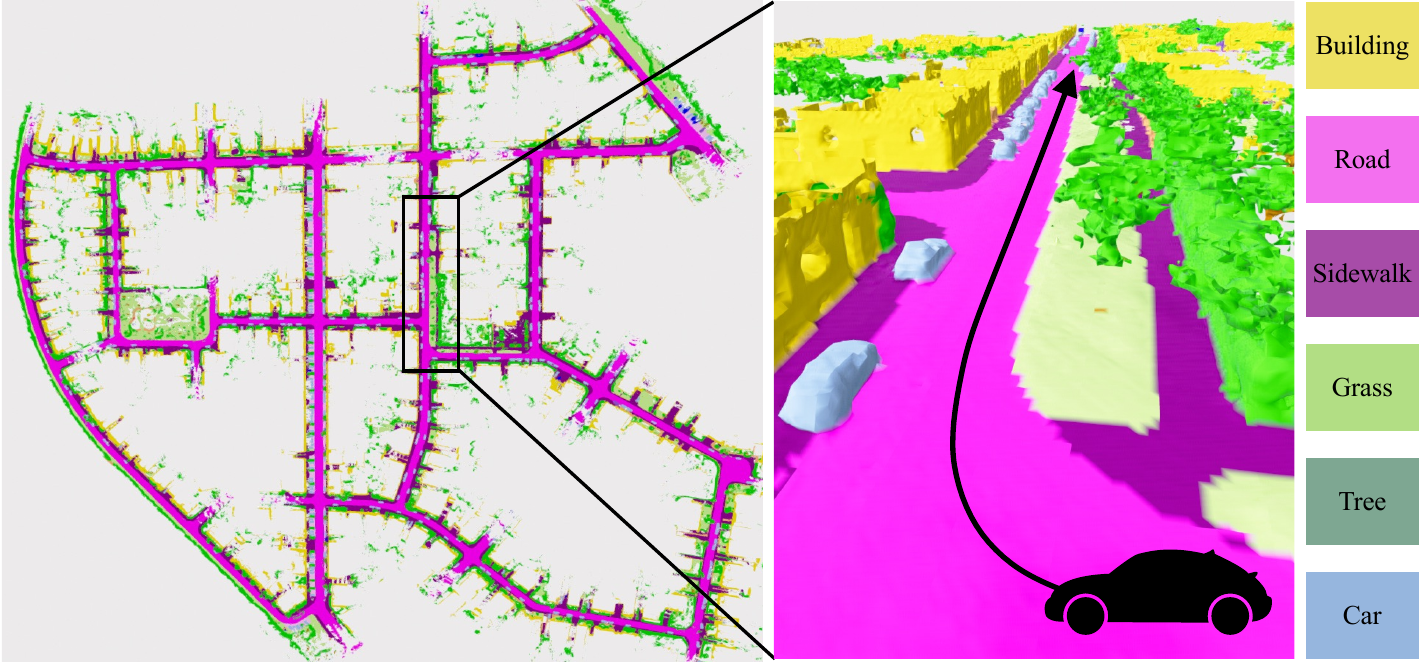}
      \caption{To successfully navigate in the complicated environment or conduct high-level or interactive tasks for a robot (such as the vehicle shown in the figure), semantic information that categorizes surrounding objects at a human-readable format is required.}
      \label{fig:cover_image_semanticmap}
      \vspace{-0.2cm}
\end{figure}
%%%%%%%%%%%%%%%%%%%%%%%%%%%%%%%%%%%%%%%%%%%%%%%%%%%%%%%%%%%%%%%%%

% Outline: 
%   Describe the importance and role of mapping
%   Describe the importance of SDF-based metric-semantic mapping, compared with occupancy-based mapping (volumetric mapping)
\subsection{Motivation}
\label{sec:motivation}
\IEEEPARstart{A}{s} the basis of localization and navigation, mapping is of growing importance in robotics.
Mapping is the process of establishing an internal representation of environments which can be operated by algorithms \cite{thrun2002robotic}.
As the widely used representation, metric maps (also referred to as ``geometric maps'') store geometry of a scene and are usually defined by positions of landmarks, distance to obstacles, or binary values to indicate free and occupied space, which are critical for robots to optimize a smooth and collision-free trajectory.
However, metric maps have difficulty in maintaining the long-term consistency since geometric features are sensitive to illumination and structural changes.
Also, metric maps have limitation in encoding human-readable information.
It is inconvenient for robots to execuate abstract human instructions (\textit{e.g.,} ``navigate to the building'' and ``follow driving rules'').

In contrast, metric-semantic mapping \cite{rosinol2021kimera} is the capability to group semantic concepts into metric maps.
The inclusion of human-labeled information facilitates many tasks such as scene abstraction \cite{rosinol2021kimera} and exploration \cite{wang2020neural}.
% place recognition \cite{li2022rinet} and
% , navigation \cite{guan2021tns}
In this paper, we focus on the autonomous navigation task of ground robots in complicated environments with abundant semantic elements.
A typical scenario is shown in Fig. \ref{fig:cover_image_semanticmap}, where many different objects such as trees and buildings appear.
It is also composed of the sidewalk that is specifically designed for pedestrains.
By incorporating human-prior knowledge, the semantic map enables the vehicle to navigate along the road, finding a path that is free of collisions and avoids intersecting with sidewalks and grasslands.
However, geometry-based traversability extraction methods often face challenges in distinguishing between roads, sidewalks, and grass due to their similar structures.
Hence, this paper aims to investigate the online metric-semantic mapping method and its potential application in navigation systems.

% Outline:
%   normal computation + distance function calculation
%   GPU-based implementation, speed up xx of traditional algorithms
%   semantic mapping
%   Terrain analysis based on mesh
%   Integration and demonstration of autonomous driving systems: localization and navigation on public and private datasets
\subsection{Challenges}
We consider that a desirable metric-semantic mapping approach should meet the following requirements:
\begin{enumerate}
      \item \textbf{Accuracy:}
            The approach should aim to construct a map that closely represents real-world environments using onboard sensor data. However, factors such as measurement noise, different view angles, and limited observations affect the quality of the map construction.
      \item \textbf{Efficiency:}
            Mapping is typically a time-consuming task, as numerous map elements need to be queried and updated based on new input. It is crucial to ensure real-time and consistent performance, especially for high-resolution or large-scale mapping applications.
      \item \textbf{Versatility:}
            The resulting metric-semantic map should be capable of supporting a wide range of applications, including but not limited to localization, path planning, and environment understanding \cite{rosinol2021kimera}.
\end{enumerate}
% \item 
% \textbf{Adaptability:} The mapping approach should be flexible enough to handle dynamic environments, where changes in the surroundings can be accommodated in a timely manner. 
% \item 
% \textbf{Robustness:} The mapping method should demonstrate resilience to various challenging scenarios, such as adverse weather conditions, poor lighting conditions, and occlusions, ensuring reliable performance under diverse circumstances. \end{itemize}

\subsection{Contributions}
As the \textbf{primary dcontribution}, we propose an online mapping system to address these challenges.
This system leverages LiDAR-visual-inertial sensing to estimate the real-time state of the robot and construct a lighweight and global metric-semantic mesh map of the environment.
To achieve this, we build upon the work of NvBlox \cite{millane2023nvblox} and thus utilize a signed distance field (SDF)-based representation.
This representation offers the advantage of constructing surfaces with sub-voxel resolution, enhancing the accuracy of the map.
While the focus of this paper is on mapping outdoor environments, the proposed solution is easily adaptable for various applications.
The modular system consists of four primary components:
\begin{enumerate}
      \item \textit{State Estimator} (Section \ref{sec:mapping_state_estimator}) is a LiDAR-visual-inertial odometry (LVIO) module implementing the Extended Kalman Filter (EKF) to estimate real-time sensors' poses with a local and sparse color point cloud.
      \item \textit{Semantic Segmentation} (Section \ref{sec:semantic_segmentation}) is a pre-trained convolutional neural network (CNN) that assigns a class label to every single pixel of each input image. A novel dataset that categorizes objects into diverse classes for the network training is also developed.
      \item \textit{Metric-Semantic Mapping} (Section \ref{sec:mapping_metric_semantic_mapping}) takes sensors' measurements and poses as input, and constructs a 3D global mesh of environments using the implicit SDF-based volumetric representation with semantic annotations from the 2D pixel-wise segmentation. The whole pipeline is implemented in parallel with the GPU and thus achieves the real-time performance. To approximate the surface geometry more accurate and complete (\textit{e.g.,} less holes), the original distance calculation is improved.
      \item \textit{Traversability Analysis} (Section \ref{sec:mapping_traversability}) identifies drivable areas by analyzing the geometric and semantic attributes of the resulting mesh map, thus narrowing the search space for subsequent motion planning.
\end{enumerate}

% As the second contribution (Section \ref{sec:navigation}), we present an autonomous navigation system that utilizes the resulting metric-semantic map in localization and motion planning algorithms.
% It has been validated with a real-world mobile robot.
% We highlight that the introduced semantic information encodes human instructions, which guides the robot to safely navigate through complex and unstructured environments.

% a traversability analysis method that identifies drivable regions by jointly analyzing geometric and semantic properties of the resulting mesh map.
% we present an autonomous navigation system that utilizes the resulting metric-semantic map in localization and motion planning algorithms.
% It has been validated with a real-world mobile robot.

The \textbf{second contribution} is an extensive experimental evaluation focusing on mapping. We evaluated the mapping system using both public datasets and our own collected data, including the \textit{SemanticKITTI} \cite{behley2021benchmark}, \textit{SemanticUSL} \cite{jiang2021lidarnet}, and \textit{FusionPortable} dataset \cite{jiao2022fusionportable}. Additionally, we collected two test sequences on campus, covering outdoor scenes with buildings, roads, and grasslands.
Our robot system utilizes maps constructed from these self-collected sequences, enabling the robot to complete point-goal navigation missions.
% The code and self-collected sequence
% To benefit the research community, we will publicly release our code, implementation details, and the self-collected dataset\footnote{\url{https://github.com/HKUSTGZ-IADC/cobra}}.

The \textbf{third contribution} encompasses real-world experiments on autonomous navigation employing the metric-semantic map created by our mapping method.
This effort effectively bridges the previously unconnected realms of semantic mapping and navigation.
% Key aspects of the work involve generating an occupancy map after terrain evaluation, a motion planning algorithm that integrates heuristics with the vehicle's nonholonomic constraints, and its incorporation into an existing navigation framework \cite{liu2021role}.
The semantic data encoded in the map translates human instructions, thereby enabling robots to navigate safely within unstructured environments.
We will publicly release the code of semantic mapping and self-collected datasets in the project website\footnote{\url{https://gogojjh.github.io/projects/2024_semantic_mapping}}\footnote{\url{https://github.com/gogojjh/cobra}}.

\begin{comment}
\subsection{Organization}
\label{sec:organization}
The rest of this paper is organized as follows.
Section \ref{sec:related_work} reviews the relevant literature.
Section \ref{sec:preliminary} provides preliminaries and basic notions of this paper.
Section \ref{sec:mapping} introduces all components of the proposed metric-semantic mapping method.
Section \ref{sec:experiment} presents experimental results.
In Section \ref{sec:discussion}, we provide discussion.
Finally, Section \ref{sec:conclusion} concludes this article.
\end{comment}

\section{Related Work}
\label{sec:related_work}

\begin{table*}[]
  \centering
  \caption{Differences in existing works on semantic mapping.}
  \renewcommand\arraystretch{1.0}
  \renewcommand\tabcolsep{8pt}
  \scriptsize
  \begin{tabular}{c|c|c|c|c|c|c}
    \Xhline{0.03cm}
    \textbf{Method}                                  & \textbf{Metric Mapping}                       & \textbf{Processing Unit} & \textbf{Semantic Update}       & \textbf{Map Representation} & \textbf{Scale} & \textbf{Application} \\
    \hline
    SLAM++ \cite{salas2013slam++}                    & KineticFusion \cite{newcombe2011kinectfusion} & GPU                      & Objects' pose optimizaiton     & TSDF                        & Indoor         & AR                   \\
    \hline
    SemanticFusion \cite{mccormac2017semanticfusion} & ElasticFusion \cite{whelan2015elasticfusion}  & GPU                      & Bayesian update                & TSDF                        & Indoor         & Not presented        \\
    \hline
    Mask-Fusion \cite{runz2018maskfusion}            & ElasticFusion \cite{whelan2015elasticfusion}  & GPU                      & Geometry enhances segmentation & TSDF                        & Indoor         & Grasping; AR         \\
    \hline
    Voxbox++ \cite{grinvald2019volumetric}           & VoxBlox \cite{oleynikova2017voxblox}          & CPU                      & Bayesian update                & TSDF                        & Indoor         & Not presented        \\
    \hline
    BKISemMapping \cite{gan2020bayesian}             & BGKOctoMap \cite{doherty2019learning}         & CPU                      & Bayesian Kernal Inference      & 3D occupancy grid           & Outdoor        & Not presented        \\
    \hline
    Kimera \cite{rosinol2021kimera}                  & VoxBlox \cite{oleynikova2017voxblox}          & CPU                      & Bayesian update                & TSDF                        & Indoor         & Scene Graph          \\
    \hline
    Sni-SLAM \cite{zhu2023sni}                       & NICE-SLAM \cite{zhu2022nice}                  & GPU                      & Optimization                   & Radiance Field              & Indoor         & Not presented        \\
    \hline
    \textbf{Ours}                                    & NvBlox \cite{millane2023nvblox}               & GPU                      & Bayesian update                & TSDF                        & Outdoor        & Navigation           \\
    \Xhline{0.03cm}
  \end{tabular}
  \label{tab:diff_semanticmap}
  \vspace{-0.30cm}
\end{table*}
%%%%%%%%%%%%%%%%%%%%%%%%%%%%%%%%%%%%%%%%%%%%%%%%%%%%%%%%%%

This section reviews the current literature on mapping and navigation techniques, focusing specifically on algorithms developed for mobile robots in unstructured environments.

%%%%%%%%%%%%%%%%%%%%%%%%%%%%%%%%%%%%%%%%%%%%%
\subsection{Geometric Mapping}
% Related works on geometric mapping commonly research on the scalability and expressivity problems.
% Scalability is often referred to computation time and memory footprint of mapping a large environments over a long period, while the expressivity is about capturing precise geometric details of surroundings.
% There exist real-time algorithms reconstructing meshes directly from RGB-D or LiDAR scans \cite{}.
Existing map representations are categorized into explicit and implicit approaches.
Explicit representations such as point clouds and surfels are widely studied in localization \cite{jiao2021robust}.
But points or surfels lack connectivity, where latent structural information is missing.
Another type of explicit representation is the triangular mesh, where structural information through vertices and triangle facets are preserved.
Meshes can depict manifold structures and topology of objects, which have been applied in scene reconstruction \cite{lin2023immesh} and planning \cite{putz2021continuous}.
However, explicit representations have difficulty in maintaining the up-to-date map over a long period \cite{schmid2022panoptic}, where environments always are changing (\textit{e.g.,} dynamic objects).

Implicit representations of environments are categorized into volumetric, elevation, and radiance field-based mappings.
The $2.5$D elevation map, efficient for legged robots' footstep planning, stores height as a Gaussian variable per grid but falls short in multi-layered scenarios and constraining $6$-DoF motions \cite{yang2022real}.
Radiance field approaches \cite{zhu2022nice} offering the ability to infer unseen areas but at the cost of high computational demands for large-scale mapping.
Volumetric methods store 3D scene geometry using discretized volumes, facilitating parallel GPU implementation for real-time applications.
Approaches include occupancy grid mapping, which assigns occupancy probabilities to voxels \cite{gan2022multitask}, and SDF-based mapping \cite{oleynikova2017voxblox}, capturing precise surface geometries with distance functions.

Our approach leverages the Truncated Signed Distance Function (TSDF) for environment representation, utilizing GPU parallelization to enhance mapping efficiency. We introduce a non-projective distance calculation to accurately estimate voxel distances, sidestepping the memory-heavy ESDF creation needed for Voxblox's collision detection. Instead, we utilize mesh-based traversability analysis and occupancy data for optimizing ground robot navigation.

% \rt{
%     Signed Distance Fields (SDFs) represent the environment as a continuous function that encodes the minimum distance to the nearest surface for each point in space, with positive values for free space and negative values for occupied space.
%     This representation provides smooth gradients and allows for efficient surface extraction, making it suitable for applications such as surface reconstruction and mesh generation.
%     Truncated Signed Distance Fields (TSDFs) extend the concept of SDFs by limiting the range of distance values to a predefined truncation threshold. This approach reduces memory requirements and enables efficient fusion of multiple depth measurements from range sensors.
% }

% Helen \textit{et al.} proposed the Voxblox.
% Alex \textit{et al.} proposed the C-blox, the submap-based mapping system, incorporate loop closure detection and pose graph optimization techniques to correct accumulated drift over time, ensuring the consistency and accuracy of the map during a large-scale and long-term mission.
% Han \textit{et al.} proposed the FIESTA to efficiently construct ESDF directly on sensor data.

\subsection{Semantic Mapping}
Semantic maps often build upon geometric representations by annotating map elements with labels.
Semantic mapping is often coupled with segmentation algorithms, such as DeepLab \cite{chen2017deeplab}, by classifying voxels into object categories.
The pioneering works in real-time metric-semantic mapping is SLAM++ \cite{salas2013slam++}, where semantic objects are represented with CAD models and their poses are optimized independently.
Recent studies such as SemanticFusion \cite{mccormac2017semanticfusion}, Mask-Fusion \cite{runz2018maskfusion}, and Voxblox++ \cite{grinvald2019volumetric} have developed dense, voxel-based semantic maps, utilizing the map's geometry to enhance frontend segmentation.
The Sni-SLAM \cite{zhu2023sni} is proposed as the NeRF semantic mapping method.
Table \ref{tab:diff_semanticmap} summarizes some of them.
As our closest work, Kimera \cite{rosinol2021kimera} leverages a CPU-based framework (built upon VoxBlox) that uses RGB-D or stereo sensing to produce dense maps and employs visual-inertial odometry for motion estimation, mainly focusing on indoor environments.
Conversely, our approach is specifically designed for the challenges of outdoor environments, introducing four major enhancements:
\textit{1)} LiDAR-visual-inertial sensing, which offers an extended measurement range, thereby substantially broadening the applicability of semantic mapping;
\textit{2)} enabling the construction of large-scale maps in real-time through the application of GPU parallelization techniques;
\textit{3)} introducing a comprehensive real-world and campus-scene semantic segmentation dataset;
\textit{4)} further leveraging the resulting metric-semantic map for localization and global planning purposes.

% Often semantic maps include a geometric representation onto which the semantic information is projected, and thus in these cases, they may contain more information than geometric maps.

% To enable context-aware interpretation of the environment, researchers also explore the methodology of semantic mapping. 

% Herb et al. \cite{herb2021lightweight} proposed a semantic mapping framework directly on meshes.
% Gan et al. \cite{gan2020bayesian} xxx.
% This work \cite{gan2022multitask} focuses on applying the proposed mapping framework
% to build a second traversability layer on top of the first
% semantic layer developed in the authors' previous work \cite{gan2020bayesian}
% Antoni \textit{et al.} proposed the Kimera, construing a hierarchical dynamic scene graph with a backbone of semantic mapping.
% Tian \textit{et al.} presented the Kimera-Multi, enabling multi-robot collborative volumetric mapping and mesh-based drift correction algorithm.

%%%%%%%%%%%%%%%%%%%%%%%%%%%%%%%%%%%%%%%%%%%%%
% \subsection{Navigation in Unstructured Environments}
\subsection{Terrain Traversability Recognition}
The difficulty of navigation in unstructured environments mainly stems from variations of terrains.
Several works \cite{miki2022elevation} obtain traversability maps by extracting geometric attributes of the surface from LiDAR, including slope, height variation, and roughness, etc.
But in many unstructured environments, path boundaries are commonly unclear and hardly inferred from geometry.
Several following works \cite{gan2020bayesian,guan2021tns,gan2022multitask} employ semantic segmentation to identify traversability by encoding prior human knowledge.
The work most similar to ours is TNS \cite{guan2021tns}, which generates a 2D traversability grid map by combining semantic and geometric information to develop an autonomous excavator application.
In comparison, our work focuses on constructing a 3D global metric-semantic map, which offers a more comprehensive and versatile representation for environments.
We use the map to benefit several tasks such as localization and motion planning.
Real-world navigation experiment with a robot is demonstrated.

\section{Preliminaries}
\label{sec:preliminary}

% \subsection{LiDAR-IMU-Visual Sensing for Mapping}
\subsection{Sensor Configuration}
% This paper utilizes the LiDAR-Visual-IMU (LVI) setting to collect data for mapping.
% These sensors have complementary advantages, making them well-suited in outdoor environments:
% \begin{enumerate}
%     \item \textit{IMU} provides high-rate measurements of linear acceleration and angular velocities, enabling smooth and high-rate motion-estimates.
%     \item \textit{LiDAR} offers 3D point clouds that directly measure structures of surrounding environments, which are insensitive to illumination- and view-changes.
%     \item \textit{Camera} captures dense RGB images and can classify objects into fine-grained categories than LiDARs. But they need external computation to recover 3D structures. While some may argue that RGB-D sensors can provide depth, their accuracy and reliability are sensitive to objects' distance and lighting conditions.
% \end{enumerate}

This paper employs a LiDAR-Visual-IMU (LVI) configuration for data collection in mapping, leveraging the complementary strengths of each sensor in outdoor environments. The \textit{IMU} offers high-rate linear acceleration and angular velocity measurements for accurate motion estimation. \textit{LiDAR} provides 3D point clouds for direct measurement of environmental structures, unaffected by changes in illumination or viewpoint. \textit{Cameras} capture dense RGB images, enabling fine-grained object classification, though they require external processing for 3D structure recovery. Despite the potential of RGB-D sensors for depth information, their performance is limited by distance and lighting variations.

% Our mapping system jointly uses the LVI setting to estimate the real-time states of the robot states while simultaneously constructing a global metric-semantic map.
% Due to LiDARs' active nature in measuring distance, point clouds enjoy high geometric consistency across consecutive frames.
% Exploiting this feature, we design a LiDAR-centric odometry to enforce the accuracy of state estimation.
% For metric mapping, we directly utilize LiDAR measurements to construct the SDF-based representations of the environment.
% Semantic segmentation is performed using camera data to estimate pixel-wise labels.
% The visual segmentation algorithms can distinguish objects such as ``road'' and ``sidewalk,'' which have similar geometric structures.
% The resulting label images are used as input by the subsequent semantic mapping module, where map elements are associated with 2D labels through projection.
% Before delving into the details of our mapping system, we introduce basic notations and describe the calibration process of multiple sensors in folloing sections.
% The complete approach is detailed in Section \ref{sec:mapping}.
% Considering different features of these sensors, 
% \subsection{Advantages of LiDAR-Inertial-Visual Sensing}

Our mapping system leverages the LVI setup to simultaneously estimate the robot's real-time states and construct a global metric-semantic map. Utilizing the active sensing capabilities of LiDAR for consistent geometric measurements across frames, we develop a LiDAR-centric odometry for precise state estimation. Metric mapping employs LiDAR data to create SDF-based environmental representations, while semantic segmentation is achieved through camera data for pixel-wise object labeling, effectively differentiating between similar structures like roads and sidewalks. These labeled images feed into a semantic mapping module, linking map elements with 2D labels via projection. We precede the detailed mapping system exposition with an introduction to basic notations and the sensor calibration process.

\subsection{Notions and Definitons}
\label{sec:notion}
In this paper, we consider the minimal LVI setting shown in Fig. \ref{fig:device-handheld}.
Frames of the world, LiDAR, IMU, and camera are defined as $()^{w}, ()^{l}$, $()^{b}$, $()^{c}$ respectively.
The IMU frame is commonly treated as the base frame.
We use $\mathbf{t}\in\mathbb{R}^{3}$ and $\mathbf{R}\in SO(3)$ to represent the 3-D translation and rotation.
Especially, the rotation matrix is from the Lie group $SO(3)$ where $\mathbf{R}^{\top}\mathbf{R}=\mathbf{I}$, $\det\mathbf{R}=1$.
% For convencience, we also represent the transformation using the $4\times 4$ transformation matrix $\mathbf{T}=(\mathbf{R},\mathbf{t})$.
With these notions, we can reprense a sensor pose like the IMU in the world frame at time $k$ as $(\mathbf{R}^{w}_{b_{k}}, \mathbf{t}^{w}_{b_{k}})$.
The basic element in volumetric mapping is the voxel.
Each voxel is represented by $V_{i}$, where $i$ denotes the index.
The size of each voxel is denoted by $\nu$.
The set of all defined semantic label is denoted by
$\mathcal{L}=\{\text{road}, \text{sidewalk}, \text{vegetation}, \cdots\}$.

\subsection{Synchronization and Calibration}
\label{sec:sensor_configuration}
% We use an Field Programmable Gate Array (FPGA) to generate an external signal trigger to synchronize clocks of all sensors.
% This guarantees data collection across multiple sensors with minimum latency.
% The FPGA receives a pulse-persecond (PPS) signal from the GPS and outputs different frequencies of signal to sensors.
% Spatial-termpoal calibration (\textit{i.e.,} intrinsics, extrinsics, and time offset) is essential to multi-sensor fusion.
% Regarding the spatial calibration, we use the Matlab calibration toolbox to calibrate intrinsics of the camera, while moving the device with significant rotation and translation to calibrate the camera-IMU extrinsics before a checkboard using Kalibr \cite{furgale2013unified}.
% To obtain the LiDAR-camera extrinsics, we propose the checkboard-based calibration method \cite{jiao2023lce}. The checkerboard is the target that provides distinctive corners and boundaries for LiDAR-camera data association. The extrinsics are optimized by minimizing point-to-plane and line-to-plane distance.

% Communication latency and time for data processing are commonly unknown during a mission, making the estimation of the time offset between a LiDAR and a camera challenging.
% Instead of estimating the offset directly, we turn to find the optimal transformation between the LiDAR frame and camera frame.
% It is described in Section \ref{sec:mapping_state_estimator}.

We employ a Field Programmable Gate Array (FPGA) to synchronize sensor clocks via an external signal trigger, ensuring minimal latency in data collection across multiple sensors. The FPGA, receiving a pulse-per-second (PPS) signal from the GPS, adjusts the signal frequencies for each sensor. Spatial-temporal calibration (\textit{i.e.,} intrinsics, extrinsics, and time offsets) is crucial for multi-sensor fusion. For spatial calibration, the Matlab calibration toolbox calibrates camera intrinsics, and we employ significant rotation and translation movements alongside Kalibr \cite{furgale2013unified} to calibrate camera-IMU extrinsics using a checkerboard. The LiDAR-camera extrinsics are determined using a checkerboard-based method \cite{jiao2023lce}, optimizing extrinsics by minimizing point-to-plane and line-to-plane distances for precise LiDAR-camera data association. Given the challenge of unknown communication latency and processing time, we optimize the transformation between LiDAR and camera frames rather than estimating time offsets directly, as detailed in Section \ref{sec:mapping_state_estimator}.

% The resulting transformation should be slightly different from the calibrated extrinsics. 
% \begin{equation}
%   \underset{\bm{R}\in SO(3),\ \bm{t}\in \mathbb{R}^{3}}{\arg\min}
%   \sum\ [\bm{n}_{i}^{\top}(\bm{g}_{i} - \bm{R}\bm{p}_{i} - \bm{t})]^{2}
% \end{equation}
\section{Mapping}
\label{sec:mapping}

%%%%%%%%%%%%%%%%%%%%%%%%%%%%%%%%%%%%%%%%%%%%%%%%%%%
\begin{figure}[t]
  \centering
  \includegraphics[width=0.475\textwidth]{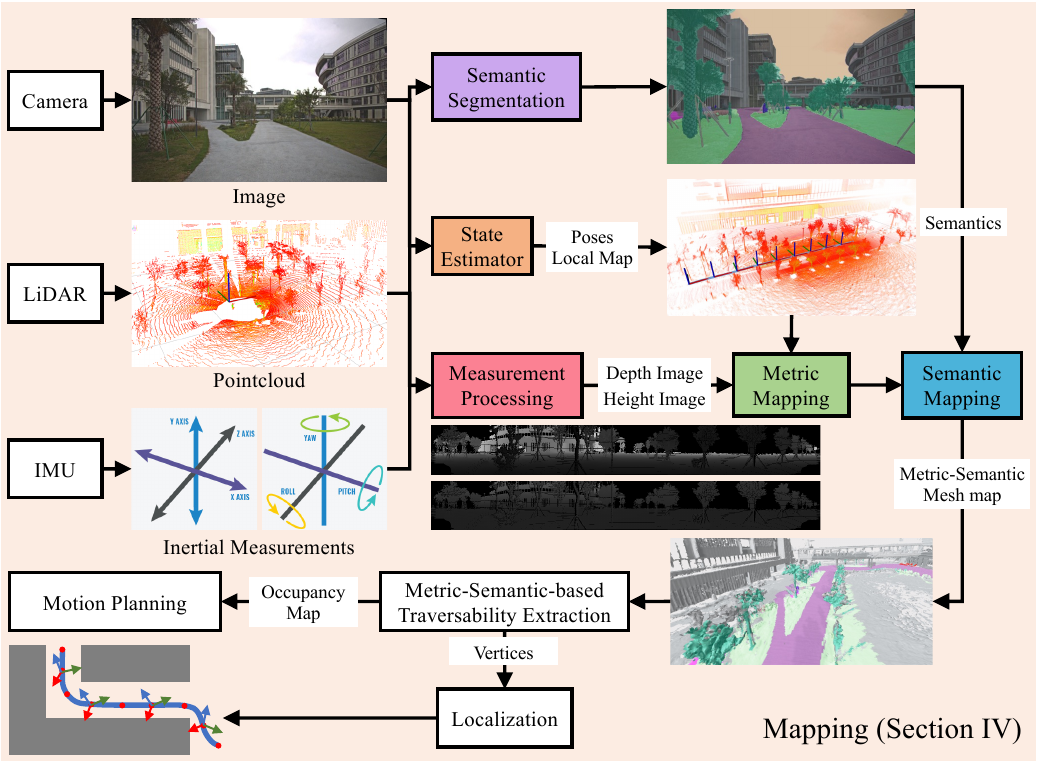}
  \caption{Block diagram illustrating the full pipeline of the proposed mapping system.
    The system starts with the state estimation (see Section \ref{sec:mapping_state_estimator}).
    The segmentation module (see Section \ref{sec:semantic_segmentation}) annotates each image pixel with a label.
    The measurement proecssing module converts point clouds into range and depth images.
    The mapping (see Section \ref{sec:mapping_metric_semantic_mapping}) constructs a global metric-semantic mesh map.
    The resulting map is extracted with traversable regions (see Section \ref{sec:mapping_traversability}), and then used for localization and generating a collision-free path by a motion planning algorithm (see Section \ref{sec:mapping_navigation}).}
  \label{fig:system_pipeline}
  \vspace{-0.35cm}
\end{figure}
%%%%%%%%%%%%%%%%%%%%%%%%%%%%%%%%%%%%%%%%%%%%%%%%%%%

Fig. \ref{fig:system_pipeline} shows the architecture of the proposed mapping system.
The system starts with a \textbf{state estimator} (see Section \ref{sec:mapping_state_estimator}), in which sensors' poses are estimated from sensor measurements.
The \textbf{semantic segmenation} module (see Section \ref{sec:semantic_segmentation}) predicts pixel-wise labels for each image.
It can be replaced by a point cloud-based segmentation network.
The \textbf{metric-semantic mapping} module (see Section \ref{sec:mapping_metric_semantic_mapping}) utilizes multi-model data (e.g., point clouds, RGB images, labeled images) to constructs the TSDF-based representation of environments.
% The marching cube algorithm \cite{newman2006survey} is applied to extract polygonal meshes.
The resulting mesh map consists of geometric and semantic information of environments.
It is further analyzed by the \textbf{traversability extraction} module (see Section \ref{sec:mapping_traversability}) to support navigation tasks, \textit{i.e.,} localization and motion planning (see Section \ref{sec:mapping_navigation}.
All these modules use ROS's ``Subscriber-Publisher'' mechanism to transfer data.

\subsection{State Estimator}
\label{sec:mapping_state_estimator}
% \rt{
%     The state estimator implements LVI odometry to provide real-time poses, which is modified from the R3LIVE system \cite{lin2022r}. R3LIVE consists of the LIO subsystem and the VIO subsystem, estimating sensors' poses and local maps in a coarse-to-fine manner.}
% The LIO subsystem first propagates high-rate motion using IMUs' measurements and then corrects the motion distortion of each LiDAR scan.
% It constructs a 3D map in the world frame, while only keeping points within a local region to to reduce memory cost.
% It then leverages the error-state iterated Kalman filter (ESIKF) to minimize the point-to-plane residuals to estimate the LiDAR's states. 

The state estimator utilizes LVI odometry for real-time pose estimation. It is adapted from the R3LIVE system \cite{lin2022r}, which integrates LIO and VIO subsystems for sensor pose and local map estimation in a coarse-to-fine approach. The LIO subsystem uses IMU measurements for high-rate motion propagation and LiDAR scans to construct a 3D map, focusing on a local region to minimize memory usage. It employs an error-state iterated Kalman filter (ESIKF) to refine LiDAR state estimates by minimizing point-to-plane residuals. The residual is formulated as
\begin{equation}
  \mathbf{0} = \mathbf{n}_{j}^{w\top}[\mathbf{R}^{w}_{l_{k}}\mathbf{p}^{l_{k}}_{j} + \mathbf{t}^{w}_{l_{k}} - \mathbf{q}^{w}_{j}],
\end{equation}
where $j$ is the index of a point in the LiDAR scan, $\mathbf{n}^{w}_{j}$ is the normal vector of the corresponding plane, and $\mathbf{q}^{w}_{j}$ is a point lying on the plane.
The subsequent VIO subsystem renders a 3D map with RGB color with input images, \textit{i.e.,} each map point is represented as $\{\mathbf{p}, \mathbf{c}=[R,G,B]^{\top}\}$.
It computes camera's pose by minimize photometric errors between frame points and corresponding map points taking $\mathbf{R}^{c}_{l}\mathbf{R}^{l_{k}}_{w}$ as an initial guess. We do not directly setting $\mathbf{R}^{c}
  _{l}\mathbf{R}^{l_{k}}_{w}$ as the camera's pose due to the existing of non-zero time offset between the camera and LiDAR. The photometric error is defined as
\begin{equation}
  \mathbf{0}
  =
  \mathbf{c}_{j}^{l_{k}}
  -
  \mathbf{I}_{k}[\kappa(\mathbf{R}^{c}_{w}\mathbf{p}^{w} + \mathbf{t}^{c}_{w})],
  \label{equ:photometric_error}
\end{equation}
where $\kappa(\cdot)$ projects a 3D point onto the image plane and $\mathbf{I}(u, v)$ returns the linearly interpolated RGB color at the pixel.
Unlike the original R3LIVE approach, we exclude RGB points older than $3$ seconds from the map for alignment, significantly reducing the memory footprint. After each frame, we relay updated sensor poses and undistorted point clouds to the following mapping modules.

% The map here is composed of a set of sparse 3D points with color and mainly serves for localization. It is different from the map described in Section \ref{sec:mapping_metric_semantic_mapping}.
% Denoting $\{\bm{p}^{l_{k}}_{j}\}$ as the LiDAR scan, we utilize the point-to-plane measurement model to constain the current pose as
% \begin{equation}
%     \bm{0} = \bm{n}_{j}^{w\top}[\bm{R}^{w}_{l_{k}}(\bm{p}^{l_{k}}_{j} + \bm{\omega}^{l}_{j}) - \bm{q}^{w}_{j}]
% \end{equation}
% Difference:
% \begin{enumerate}
%     \item Undistortion on raw point clouds
%     \item Undistortion on RGB images
%     \item Extract features from the ouster point cloud
%     \item Local map
% \end{enumerate}

\subsection{Semantic Segmentation}
\label{sec:semantic_segmentation}

% Confidence-aware semantic segmentation with prototype learning is utilized to output pixel-wise semantic prediction and predictive uncertainty, composed of a off-the-shelf segmentation backbone, a segmentation head and a confidence head. Specifically, each semantic class is represented as a non-learnable prototype vector, which is updated by online clustering of the data points using Optimal Transport \cite{zhou2022rethinking}. Inspired by \cite{liu2022modeling}, to guide the network to pay more attention to the areas where the predictions are uncertain, aleatoric uncertainty is estimated with the original image and model predictions being input, under the supervision of semantic ground-truth. 

We design a network that is coupled with prototype learning for segmentation. It is composed of an off-the-shelf segmentation backbone \cite{wang2020deep}, a customized segmentation head, and a confidence head.
% In the segmentation head, each class is represented by a non-learnble prototype which is updated by online clustering of the data points using Optimal Transport \cite{sacha2023protoseg}. 
To guide the network to pay more attention to the areas where the predictions are uncertain, the confidence head is used to predict pixel-wise aleatoric uncertainty \cite{kendall2017uncertainties} from images.
The detail of the network is explained in \cite{liu2022modeling}.

\subsection{TSDF-Based Volumetric Mapping}
\label{sec:mapping_metric_semantic_mapping}

% \rt{TSDF: as they are fast to construct, filter out sensor noise, and can create human-readable meshes with sub-voxel resolution. In contrast to ESDFs, they use projective distance, which is the distance along the sensor ray to the measured surface, and calculate these distances only within a short truncation radius around the surface boundary.}

After obtaining the undistorted 3D scans (\textit{e.g.,} LiDAR/ RGB-D scans) and labeled images with associated poses, our metric-semantic mapping approach incrementally builds a dense 3D map.
In traditional CPU-based serial pipeline, the time for updating voxels' values is linear to the number of data. This weakness limits the usage of sensors with large field of view (FOV) and dense measurements such as LiDARs. The state-of-the-art method (\textit{i.e.,} Voxblox \cite{oleynikova2017voxblox}) achieves the nearly real-time performance with point average.
But this manner may unavoidably cause information loss.
In contrast, the mapping pipeline including the retrieval and operation of all visible voxels is done in parallel within a GPU.
Ths pipeline consists of three key modules: \textbf{measurement preprocessing} on point clouds, \textbf{metric mapping}, and \textbf{semantic mapping}.
Here we introduce the details.

% we use a dynamically sized map that makes use of the voxel hashing approach of Niessner et al. [12]. Each type of voxel (TSDF or ESDF) has its own layer, and each layer contains independent blocks that are indexed by their position in the map.

\subsubsection{Measurement Preprocessing}
As commonly done in learning-based approaches \cite{chen2021moving}, point clouds are often converted into images and then processed in GPUs.
With the known specifications of a LiDAR (\textit{i.e.,} horizontal and vertical angular resolution $\Delta \phi$ and $\Delta \theta$ as well as the starting vertical angle $\theta_{0}$), we project an undistorted point cloud onto a depth image $D$ and height image $H$.
% Fig. \ref{fig:lidar_image_conversion} visualizes this process.
Such images are very lightweight ($100$KB \textit{v.s.} $10$MB).
Each pixel from $D$ and $H$ is calculated by the corresponding point as follows:
\begin{equation}
  \begin{aligned}
     & D( u, v) = Fr                                    ,
    \ \ \ \ \ \ \ \ \ \ \ \ H(u, v) = F(z + O),                \\
     & u       = \frac{\pi - \arctan(y, x)}{\Delta\phi} ,\ \ \
    v = \frac{\arccos(z/r) - \theta_{0}}{\Delta\theta},
  \end{aligned}
\end{equation}
where $(x, y, z)$ is the point's coordinate, $z$ is its Euclidean distance to the origin,
and $F$ and $O$ are two scalars.

% With knowing the vertical scanning angles of a LiDAR, we can project the raw point cloud onto a range image without data loss. In the image, each valid point is represented by a pixel. The pixel value records the Euclidean distance from a point to the origin. We apply the segmentation method proposed in [68] to group pixels into many clusters. We assume that two neighboring points in the horizontal or vertical direction belong to the same object if their connected line is roughly perpendicular (> 60◦) to the laser beam. We employ the breadth-first search algorithm to traverse all pixels, ensuring a constant time complexity. We discard small clusters since they may offer unreliable constraints in optimization.

\subsubsection{Metric Mapping}
\label{sec:mapping_metric}
The volumetric mapping divides the space into a set of voxels $V_{i}$.
Each voxel has a unique global coordinate $\mathbf{v}_{i}\in\mathbb{Z}^{3}$, from which the raw coordinate of its center is $\mathbf{x}_{i} = \nu\mathbf{v}_{i}=[x, y, z]^{\top}$ ($\nu$ is the voxel size).
Voxels are stored using a two-level hierarchy approach \cite{niessner2013real}.
The first level implements a hash table that maps 3D grid indices to \textit{VoxelBlocks}.
This hash table can be queried in GPU kernels using an interface based on stdgpu \cite{stotko2019stdgpu}.
Each VoxelBlock contains a small group of densely allocated $8\times 8 \times 8$ voxels which are stored contiguously in GPU memory.
In the second layer, each voxel insided in the block can be accessed.

% The proposed volumetric mapping divides the space into a set of voxels $V_{i}$ that are ordered according to their coordinates.
% Each voxel has a unique global coordinate $\mathbf{v}_{i}\in\mathbb{Z}^{3}$, from which the raw coordinate of its center is $\nu\mathbf{v}_{i}=[x, y, z]^{\top}$.
% To enable the dynamic insertion and deletion, voxels are managed and queried using the hashing approach \cite{niessner2013real}.

\begin{figure}[t]
  \centering
  \subfigure[Flat surface]
  {\label{fig:non_projective_distance_flat}\centering\includegraphics[width=0.17\textwidth]{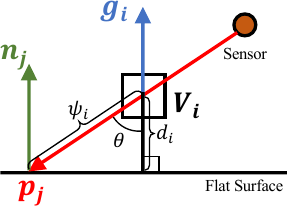}}
  \ \ \ \ \ \
  \subfigure[Curved surface]
  {\label{fig:non_projective_distance_curve}\centering\includegraphics[width=0.13\textwidth]{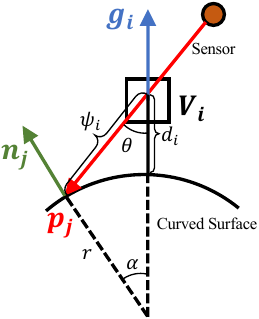}}
  \caption{The non-projective distance uses the local planarity of surfaces to approximate the true distance $d_i$.
    $\psi_{i}$ is the projective distance of the voxel.
    The gradient vector is computed as the weighted average of normal vectors.
    % $\alpha$ is the angle between the gradient and the normal, indicating the curvature of surface.
    % And $\theta$ is the angle between the gradient and the ray.
    $d_{i}$ is calculated according to equ.\eqref{equ:distance}.
    The radius of the curved surface (approximate to a circle) in (b) is marked as $r$.
  }
  \label{fig:non_projective_distance}
  \vspace{-0.3cm}
\end{figure}

In TSDF-based mapping, each voxel stores a truncated signed distance $D_{i}$, a weight $W_{i}$ to indicate the confidence, and a normalized gradient vector $\mathbf{g}_{i}\in\mathbb{R}^{3}$ of the signed distance.
Both VoxBlox and NvBlox define $D_{i}$ as the projective distance that is equal to the distance along the sensor ray to the measured surface of each voxel.
$D_{i}<0$ means that the voxel is behind the surface.
% , which is approximate to the true distance.
% In the original NvBlox implementation, each voxel stores the projective distance (\textit{i.e.,} the distance along the sensor ray to the measured surface) and the weight (\textit{i.e.,} describe the confidence).
Instead, we utilize non-projective distance, as described in \cite{pan2022voxfield}, by leveraging normal and gradient vectors to characterize the local planarity of surfaces and provide an approximation of the true distance.
Fig. \ref{fig:non_projective_distance} visualizes the non-projective distance $d_{i}$.
For each incoming depth and height image, we first compute the normal image $\mathbf{N}$.
The normal of each pixel is computed as
$\mathbf{N}(u,v) =
  \frac{(\mathbf{p}_{j,1} - \mathbf{p}_{j})\times (\mathbf{p}_{j,2} - \mathbf{p}_{j})}
  {\|(\mathbf{p}_{j,1} - \mathbf{p}_{j})\times (\mathbf{p}_{j,2} - \mathbf{p}_{j})\|}$,
% \begin{equation}
%   \mathbf{N}(u,v) =
%   \frac{(\mathbf{p}_{j,1} - \mathbf{p}_{j})\times (\mathbf{p}_{j,2} - \mathbf{p}_{j})}
%   {\|(\mathbf{p}_{j,1} - \mathbf{p}_{j})\times (\mathbf{p}_{j,2} - \mathbf{p}_{j})\|},
% \end{equation}
Where $\mathbf{p}_{j,1}$ and $\mathbf{p}_{j,2}$ correspond to points back-projected from $D(u,v-1)$ and $D(u-1,v)$, respectively. We traverse and ray-cast each pixel to retrieve visible voxels.
% Each voxel is projected onto the image, $\kappa[\mathbf{R}^{c}_{w}\mathbf{x}{i} + \mathbf{t}^{c}_{w}]$, to determine depth. 
The non-projective signed distance is then defined as
\begin{equation}
  d_{i} =
  \begin{cases}
    |\cos\theta| \psi_{i},                                                & \text{if } \alpha\approx 0 \\
    |\frac{(\cos\alpha - 1)\sin\theta}{\sin\alpha}|+|\cos\theta| \psi_{i} & \text{otherwise}
  \end{cases},
  \label{equ:distance}
\end{equation}
where $\theta$ is the angle between the ray and the gradient $\mathbf{g}_{i}$ and $\alpha$ is the angle between the gradient and the surface normal corresponding to $\mathbf{p}_{j}$.
We define $\tau$ as the truncated distance and $W_{i}=\frac{\psi+\tau}{2\tau}$ as the linear weight.
% \begin{equation}
%     W_{i}
%     =
%     W_{\text{linear}} W_{\text{dropoff}}
%     =
%     \frac{\psi+\tau}{2\tau}
%     \frac{\psi+\tau}{0.66\tau}
% \end{equation}
% float epsilon = trunc * 0.5f;
% float a = 0.5f * trunc / (trunc - epsilon);
% float b = 0.5f / (trunc - epsilon);
% if (sdf < epsilon) return 1.0f;
% if ((epsilon <= sdf) && (sdf <= trunc)) return (a - b * sdf + 0.5f);
% return 0.5f;
The distance $D_{i,k}$ and weight $W_{i,k}$ of $V_{i}$ are updated at the $k^{th}$ input data as follows
\begin{equation}
  \begin{aligned}
    D_{i,k+1}      &
    = \frac{W_{k}D_{k} + W_{i}\Gamma(d_{i},\tau)}{W_{i,k} + W_{i}}, \ \ \
    W_{i,k+1}
    = W_{i,k}+W_{i}, \\
    \Gamma(d,\tau) &
    =
    \begin{cases}
      \min(d, \tau)  & \text{if } d \geq 0 \\
      \max(d, -\tau) & \text{if } d < 0    \\
    \end{cases}
  \end{aligned}.
\end{equation}

% But this method prones to overestimate the actual Euclidean distance to the nearest surface in cases where the sensor ray is not perpendicular to the surface locally \cite{}.
% Therefore, we implement the non-projective method presented in \cite{}, where the local planar information of the surface is used, to better approximate the true distance of voxels.
% In Table \ref{}, we compared different distance and weighting methods in mapping.
% For the detailed implementations of the method, we refer our readers to this work \cite{}.

\subsubsection{Semantic Mapping}
\label{sec:mapping_semantic}
% To project the traversability map onto the image plane, we use each pixel’s depth value to retrieve the corresponding traversability score from the map and associate it with that pixel. In this way, we generate an image where each pixel has a traversability score, except those with invalid depth values. We then set a threshold for the projected score to get a binary label. An example of the projected traversability labels is given in. \rt{give equation}.

% \bt{TODO by Jianhao: provide more details}

Given the $k^{th}$ label image $\mathbf{I}_{k}$ and the associate pose, our semantic mapping only retrieve and update visible and valid voxels ($W>0$) within the camera frustum by raycasting.
Similar to the metric mapping, each voxel is projected onto the image plane to obtain the corresponding semantic label.
Different from the metric mapping, we propose that each voxel stores a discrete probability distribution, $P(l_{s})$ over the set of class labels, $l_{s}\in\mathcal{L}$.
Each new voxel is initialized with a uniform distribution over the semantic classes, as we begin with no a priori evidence as to its latent classification.
Besides labels, the segmentation network in Section \ref{sec:semantic_segmentation} also outputs per-pixel probability image $\mathbf{O}_{(u,v)}$ over the class labels $P(\mathbf{O}_{(u,v)}=l_{s}|\mathbf{I}_{k})$.
We can update the probability distribution of the $i^{th}$ voxel by means of a recursive Bayesian update \cite{rosinol2020kimera}:
\begin{equation}
  \label{equ:bayesian_filter}
  P(l_{s}|\mathbf{I}_{1,\dots,k})
  =
  \frac{1}{Z}
  P(l_{s}|\mathbf{I}_{1,\dots,k-1})
  P(\mathbf{O}_{(u,v)}=l_{s}|\mathbf{I}_{k}),
\end{equation}
where $Z$ is a constant.
The segmentation network may produce incorrect labels, while \eqref{equ:bayesian_filter} associates label hypotheses from multiple images and combine evidence iteratively.
After that, the global metric-semantic mesh is extracted using the marching cubes algorithm \cite{newman2006survey}, where the label of each vertex is extracted from the one with the highest probability.
% Our implementation accelerats the semantic mapping in GPUs by up to $10ms$.

\subsection{Traversability Extraction}
\label{sec:mapping_traversability}

% The complexity of navigation in unstructured environments mainly steams from the variation of roads.
% Taking the campus as an example (shown in Fig. \ref{fig:semantic-map-sequence00}), roads that are traversable for robots should cover streets, sidewalks, and bicycle lanes. For robots that have off-road capability such as quadruped robots, terrain is also the traversable road.
% In practical applications, the definition of traversability should consider both robots' mobility and human's order.
% The former factor can be formulated according to robots' kinodynamic property, while the latter factor is hard to be represented. It commonly requires lots of efforts in parameter turning.
% In this section, we describe the proposed traversability extraction algorithm that jointly consider both geometric and semantic properties from the resulting metric-semantic mesh map.

Traditional methods are either based on visual features \cite{furgale2010visual} or geometric structures \cite{yang2022far}, having limitations in complex unstructured environments with many road variations.
% Fig. \ref{fig:cover_image_semanticmap} shows an example.
The detection of traversability should consider both robots' mobility and human instructions.
Robots' mobility is typically formulated according to their kinomatic properties.
% For robots that have off-road capability such as quadruped robots, grassland should be traversable.
Regarding the latter factor, the introduced semantic information that encodes human knowledge benefits two aspects: identifying untraversable terrains and guiding robots to follow basic driving rules.
Therefore, this section proposes a traversability extraction method that jointly considers  geometric and semantic information from the resulting map $\mathcal{M}^{w}$.

% \subsubsection{Structure of Metric-Semantic Mesh}
% The global mesh is represented as the polygon mesh $\mathcal{M}^{w}=(\mathcal{V}, \mathcal{E}, \mathcal{F}, \mathcal{L})$, that is a collection of vertices $\mathcal{V}$, edges $\mathcal{E}$, faces $\mathcal{F}$, and labels $\mathcal{L}$.
% A set of 3D vertex 
% $\mathcal{V}$ is the set of 3D vertices. Each vertex is a 3D point: $\bm{v}_{i}=[x, y, z]^{\top}$.

% \begin{enumerate}
%     \item $\mathcal{V}$ is the set of 3D vertices. Each vertex is a 3D point: $\bm{v}_{i}=[x, y, z]^{\top}$.
%     \item $\mathcal{E}$ is the set of edges. Each edge is a line segment connecting two vertices and can be represented as $e_{i,j} = (\bm{v}_{i}, \bm{v}_{j})$ where $\bm{v}_{i}, \bm{v}_{j} \in \mathcal{V}$.
%     \item $\mathcal{F}$ is a set of faces. Each face is a polygon enclosed by edges and can be represented as $f = (e_j, e_k, e_l)$ where $e_j, e_k, e_{l} \in \mathcal{E}$.
%     \item $\mathcal{L}$ is a set of labels with the number equal to $\mathcal{V}$. Each label is a scalar $l\in\{0, 1, \dots, N\}$.
% \end{enumerate}

% The introduction of structure of meshes help us to explain the subsequent sections.
% By jointly considering both geometric and semantic properties, we can extract traversable region by checking whether all geometric properties are larger than a thresholds repectively and the semantic property is allowed to be drived.

\subsubsection{Analysis of Geometric Properties}
The 3D mesh map, represented as the polygon mesh is a collection of vertices, edges, faces, and labels \cite{lin2023immesh}.
Each face provides normal information that is suitable for terrain assessment.
We analyze the below geometric properties to determine whether the road is traversable or not from the geometric perspective: height difference $v_{hd}$, steepness $v_{s}$, and roughness $v_{r}$.
Fig. \ref{fig:mesh-terrain-sequence} visualizes some examples.
The ``height difference'' and ``steepness'' are used to indicate the risk of collision.
And the ``steepness'' indicates the changing height of terrain.
A vertex is selected if its $v_{hd}$, $v_{s}$, and $v_{r}$ are all larger than thresholds
$t_{hd}, t_{v}, t_{r}$.
After that, we get the filtered mesh map $\mathcal{M}^{w'}$.
\begin{itemize}
  \item \textbf{Height Difference} refers to the maximum difference in elevation between two points within a local region (\textit{i.e.,} a ball $\mathcal{B}$ with radius $r$): $v_{hd} = \underset{\bm{v}_{i}, \bm{v}_{j}\in\mathcal{B}}{\arg\max}\ \|\bm{v}_{i}-\bm{v}_{j}\|.$
        % \begin{equation}
        %     v_{hd} =
        %     \underset{\bm{v}_{i}, \bm{v}_{j}\in\mathcal{B}}{\arg\max}\ \|\bm{v}_{i}-\bm{v}_{j}\|.
        % \end{equation}
  \item \textbf{Steepness} refers to the degree of incline of a surface: $v_{s} = \arccos(\bm{n}_{\bm{v}_{i}}).$
        % \begin{equation}
        %     v_{s} = \arccos(\bm{n}_{\bm{v}_{i}}).
        % \end{equation}
  \item \textbf{Roughness} refers to the irregularities and unevenness of a ground: $v_{r} = \frac{1}{|\mathcal{B}|}\sum_{\bm{v}\in\mathcal{B}}\bm{n}_{\bm{v}}.$
        % \begin{equation}
        %     v_{r} = \frac{1}{|\mathcal{B}|}
        %     \sum_{\bm{v}\in\mathcal{B}}\bm{n}_{\bm{v}}.
        % \end{equation}
\end{itemize}

% Height Difference: This generally refers to the vertical distance between two points. In geographical terms, it can refer to the difference in elevation between two points on a landscape. In data analysis, it might refer to the difference between the highest and lowest values in a dataset.

% Steepness: Also known as slope or gradient, steepness refers to the degree of incline of a surface. In geographical terms, it could refer to how sharply a mountain side rises. In mathematical terms, it refers to the rate at which a line rises or falls for a given set of coordinates: steepness = vertical change / horizontal change (also known as "rise over run").

% Roughness: This can have multiple meanings depending on context. Generally, it refers to the irregularities and unevenness of a surface. In geography, it can refer to the ruggedness of a terrain, which would be determined by variations in its height over short distances. In physics, it could refer to the texture of a material at a microscopic level. In data analysis, it could refer to the variability in a dataset.

\subsubsection{Analysis of Semantic Properties}
Due to the limited FoV of cameras, several vertices in the resulting map $\mathcal{M}^{w'}$ may not be labeled and are removed.
The labeled vertices indicate the categories or classes of objects in the environment.
In our approach, we can establish a strict definition of traversability based on prior knowledge and specific requirements for a particular robot.
% For instance, we can define that regions labeled as ``road'' are considered drivable for vehicles (as depicted in Fig.\ref{fig:device-car}) and regions labeled as ``sidewalk'' or ``grass'' are not driverable.
For example, we can classify ``road'' regions as drivable for vehicles (as shown in Fig. \ref{fig:device-car}), while ``sidewalk'' or ``grass'' regions are not.

\subsection{Localization and Motion Planning}
\label{sec:mapping_navigation}
The resulting map plays a critical role in subsequent navigation tasks, serving as the global map for localization and planning.
We extract vertices from $\mathcal{M}^{w}$ to form a global point cloud map.
We use the prior map-based localization method \cite{hu2023paloc} to obtain the real-time global pose of the vehicle by registering the map of the current scan.
For motion planning, we project the vertices of the above traversable map onto a 2D occupancy grid map. Each grid cell is drivable if its associated ``occupancy probability'' is zero.
To compute a collision-free and optimal global trajectory from an initial point to a specified goal, we utilize the search-based hybrid A* algorithm.
This algorithm incorporates heuristics while accommodating the vehicle's nonholonomic constraints, enabling the generation of viable and smooth trajectories.
A series of equidistant waypoints is discretized from the resultant path and then taken by the vehicle's speed controller.
% which is then used to guide the vehicle through a lateral trajectory tracking controller (for steering adjustments) and a longitudinal speed controller (for speed changes).}

\section{Experiment}
\label{sec:experiment}

We perform the mapping experiments on both public and self-collected datasets.
First, for benchmarking, we perform experiments on three public datasets: the \textit{SemanticKITTI}, the \textit{SemanticUSL}, and the \textit{FusionPortable} dataset.
Second, we validate the proposed traversability extraction and motion planning method, with the demonstration of real-world navigation tests on an autonomous vehicle.
% Second, we test the proposed navigation systems in a real-world scenario.

% We evaluate the mapping approach on three public datasets: \textit{SemanticKITTI}, \textit{SemanticUSL}, and \textit{FusionPortable} dataset.
% And we then test the proposed traversability extraction method in real-world scenario.

% We perform simulated and real-world experiments on three platforms to test the performance of M-LOAM. First, we calibrate multi-LiDAR systems on all the presented platforms. The proposed algorithm is compared with SOTA methods, and two evaluation metrics are introduced. Second, we demonstrate the SLAM performance of M-LOAM in various scenarios covering indoor environments and outdoor urban roads. Moreover, to evaluate the sensibility of M-LOAM against extrinsic error, we test it on the handheld device and vehicle under different levels of extrinsic perturbation. Finally, we provide a study to comprehensively evaluate M-LOAM’s performance and computation time with different LiDAR combinations.

%%%%%%%%%%%%%%%%%%%%%%%%%%%%%%%%%%%%%%%%%%%%%%%%%%%%%%%%%%%%%%%%%%
\begin{figure}[t]
  \centering
  \subfigure[Mapping device]
  {\label{fig:device-handheld}\centering\includegraphics[width=0.2\textwidth]{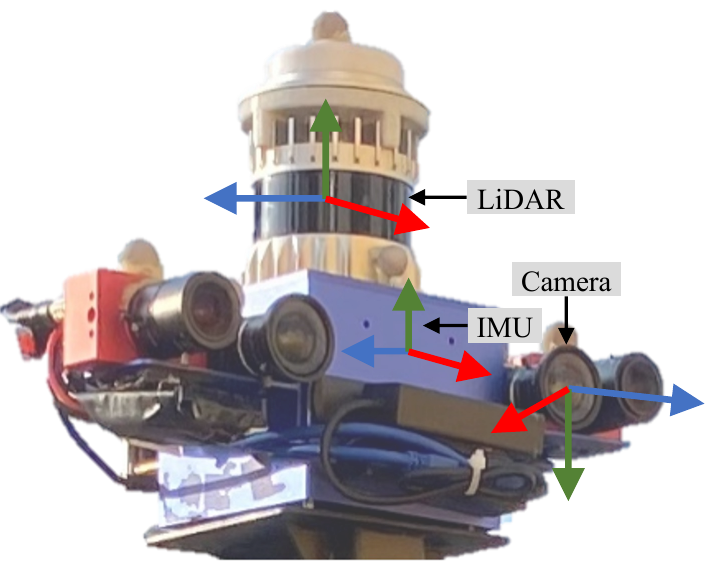}}
  \subfigure[Autonomous vehicle]
  {\label{fig:device-car}\centering\includegraphics[width=0.2\textwidth]{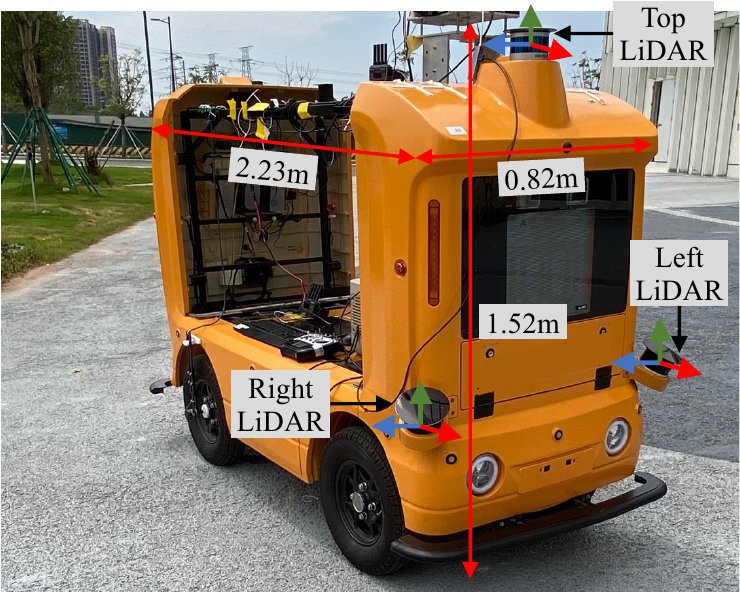}}
  \caption{(a) The mapping device that consists of a high-resolution LiDAR and camera is used to collect data for the environmental mapping. (b) The real-world vehicle provides a platform for testing the navigation system.}
  \label{fig:device}
  \vspace{-0.3cm}
\end{figure}
\begin{table}[t]
  \centering
  \caption{Parameters in experiments.}
  \renewcommand\arraystretch{1.5}
  \renewcommand\tabcolsep{4.0pt}
  \scriptsize
  \begin{tabular}{c|cc|cc}
    \Xhline{0.03cm}
    \multirow{2}{*}{Name} & \multicolumn{2}{c|}{Mapping} & \multicolumn{2}{c}{Traversability Extraction}                                                   \\ \cline{2-5}
                          & Voxel size $\nu$             & Truncated dis. $\tau$                         & Radius $r$ & Thresholds $t_{hd}, t_{v}, t_{r}$  \\
    \Xhline{0.01cm}
    Value                 & $0.25m$ or $0.3m$            & $5\nu$                                        & $0.25m$    & $0.6m, \ 20^{\circ}, \ 30^{\circ}$ \\
    \Xhline{0.03cm}
    \multicolumn{5}{l}{
      \textit{SemanticKITTI} $00, 02, 08$: $\nu=0.3m$. Others: $\nu=0.25m$.}
  \end{tabular}
  \label{tab:parameter}
  \vspace{-0.3cm}
\end{table}
%%%%%%%%%%%%%%%%%%%%%%%%%%%%%%%%%%%%%%%%%%%%%%%%%%%%%%%%%%%%%%%%%%

%%%%%%%%%%%%%%%%%%%%%%%%%%%%%%%%%%%%%%%%%%%%%%%%%%%%%%%%%%%%%%%%%
\begin{figure}[t]
  \centering
  \includegraphics[width=0.45\textwidth]{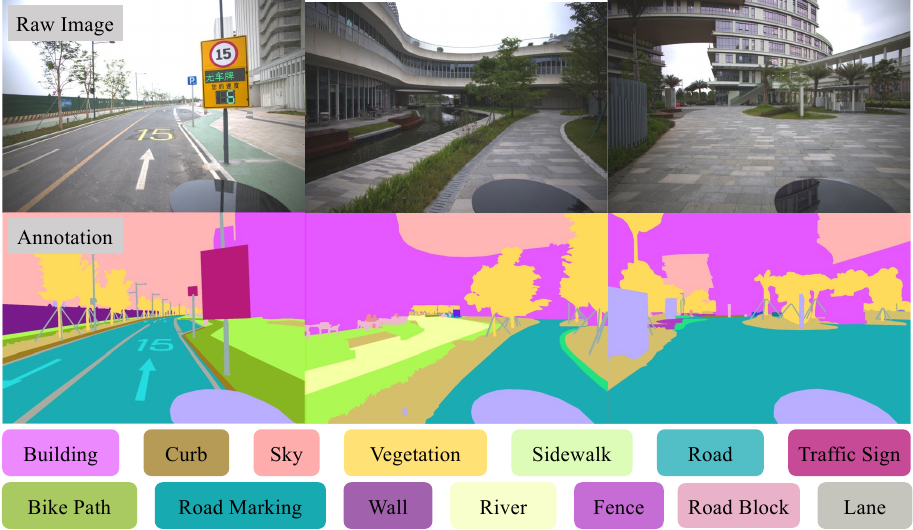}
  \caption{We show a few samples from our dataset (top) and corresponding annotations (bottom). All images are collected at the campus.}
  \label{fig:exp_annotation_image}
  \vspace{-0.35cm}
\end{figure}

\subsection{Implementation Details}
% We use the PCL library [71] to process point clouds and the
% Ceres Solver [73] to solve nonlinear least-squares problems. 
The mapping system is mainly implemented with C++ with CUDA, while the semantic segmentation is implemented in Python with the Pytorch library.
The mapping algorithms are tested on two computing platforms: a desktop PC equipped with an Intel i$9$-$12900$KF CPU, $64$GB of RAM, and an Nvidia GeForce RTX $3080$Ti GPU, as well as an embedded Nvidia Jetson ORIN $32$GB computer.
Besides public datasets, we also collect real-world data to test our mapping method.
We use a handheld multi-sensor device (see Fig. \ref{fig:device-handheld}) to collect data.
The device is installed a OS$1$ LiDAR (resolution: $128\times 1024$), two FILR BFS-U$3$-$31$S$4$C global-shutter color cameras, and one STIM$300$IMU. During the data collection, the average movement speed is around $5m/s$.
In real-world navigation experiments, we use the autonomous vehicle \cite{liu2021role} (see Fig. \ref{fig:device-car}) for tests.
The vehicle is mounted with four $16$-beam LiDARs and one Livox mid-$70$ LiDAR.
% , at the top, front, left, right, and rear positions, respectively. 
% Its average movement speed is around $3m/s$.
Table \ref{tab:parameter} shows parameters that are empirically set in experiments.
The voxel size is different in \textit{SemanticKITTI} $00,02,08$ since the scope is very large and GPU memory is limited to store all voxels.

\subsection{Dataset}

%%%%%%%%%%%%%%%%%%%%%%%%%%%%%%%%%%%%%%%%%%%%%%%%%%%%%%%%%%%%%%%%%
% \begin{table}[t]
%   \centering
%   \caption{Segmentation Accuracy}
%   \renewcommand\arraystretch{1.0}
%   \renewcommand\tabcolsep{12pt}
%   \begin{tabular}{cccc}
%     \toprule
%     Methods & metric1 & metric2   & metric3 \\
%     \midrule[0.03cm]
%     Road    & xx      & Easy      & $10\%$  \\
%     Terrain & xx      & Forbidden & $10\%$  \\
%     \bottomrule
%   \end{tabular}
%   \label{tab:exp_semantation}
% \end{table}
%%%%%%%%%%%%%%%%%%%%%%%%%%%%%%%%%%%%%%%%%%%%%%%%%%%%%%%%%%%%%%%%%

This section presents the segmentation dataset that is collected at the campus and used to train our semantic segmentation network.
We collect $15$ data sequences covering most outdoor places of the campus and annotate $1092$ images of size $2048 \times 1536$.
We split $95\%$ and $5\%$ images into the train and validation set, respectively.
%  with $23$ classes\footnote{.
% Table \ref{tab:exp_annotation_class} shows definitions of involved classes.
Unlike existing datasets \cite{cordts2016cityscapes} that focus on urban areas, our dataset consists of many types of terrains (see Fig. \ref{fig:exp_annotation_image}) and anomaly objects, which is beneficial to downstream tasks including planning and navigation of ground robots.
The network is pretrained on both the Cityscales \cite{cordts2016cityscapes} and our dataset, obtaining the $54.53\%$ \textit{Mean Intersection Over Union (mIoU)} on the validation set.

\subsection{Metric-Semantic Mapping Experiments}
\subsubsection{Evaluation Metrics}
We extract vertices from the reconstructed metric-semantic mesh map produced by our method for evaluation.
The set of vertices form a point cloud $\mathcal{M}$ that is compared with respect to the ground-truth point cloud $\mathcal{G}$ based on five metrics: \textit{Reconstruction Error (RE)} in terms of the RMSE, \textit{Chamfer Distance (CD)}, \textit{Reconstruction Coverage (RC)}, \textit{mIoU}, and \textit{Accuracy of Correctly Labeled Points (Acc)}.
The latter two metrics evaluate the quality semantic segmentation of the resulting map \cite{behley2021benchmark}.
% \textit{Semantic Mapping Metric} is calculated in terms of the \textit{Mean Intersection Over Union (mIoU) and Accuracy of Correctly Labeled Points (Acc)} \cite{behley2021benchmark}.
% The latter metrics are designed for semantic mapping.
% They are computed as follows:
% These metrics are explained in Algorithm \ref{alg:mapping_evaluation_metric}.
\begin{itemize}
  \item \textit{Reconstruction Error} computes the average point-to-point distance between $\mathcal{M}$ and $\mathcal{G}$ \cite{pan2022voxfield}:
        \begin{equation}
          \begin{aligned}
            \textit{RE}
             & =
            \sqrt{
              \frac{1}{|\mathcal{M}|}
              \sum_{\bm{p}\in\mathcal{M}}
              {\underbrace{\min(2\nu,\ \|\bm{p} - \bm{q}\|)}_{d(\bm{p}, \mathcal{G})}}^{2}
            },
          \end{aligned}
        \end{equation}
        where $\nu$ is the size of a voxel and $\bm{q} \in \mathcal{G}$ is the nearest point to $\bm{p}$.
  \item \textit{Chamfer Distance} computes the Chamfer-L1 Distance \cite{mescheder2019occupancy} as:
        \begin{equation}
          \textit{CD}
          =
          \frac{1}{2|\mathcal{M}|}
          \sum_{\bm{p}\in\mathcal{M}}
          d(\bm{p}, \mathcal{G})
          +
          \frac{1}{2|\mathcal{G}|}
          \sum_{\bm{q}\in\mathcal{G}}
          d(\bm{q}, \mathcal{M}).
        \end{equation}
  \item \textit{Reconstruction Coverage} is defined as the ratio between the number of GT points that do have a nearby point from $\mathcal{M}$ ($\leq 2\nu$) and the point number of $\mathcal{G}$ \cite{pan2022voxfield}.
        % \begin{equation}
        %   \textit{RC} =
        %   % \frac{N_{inlier}}{|\mathcal{M}_{gt}|},
        %   N_{inlier}/|\mathcal{M}_{gt}|,
        % \end{equation}
        % where $N_{inlier}$ is the number of points of the ground-truth point cloud $\mathcal{M}_{gt}$ which have the nearest point ($\|\bm{p}-\bm{q}\|<d_{inlier}$ ) from the estimated point cloud $\mathcal{M}_{est}$.
  \item \textit{Semantic Mapping Score} is calculated in terms of the \textit{Mean Intersection Over Union (mIoU) and Accuracy of Correctly Labeled Points (Acc)} \cite{behley2021benchmark}.
        %       as:
        %       \begin{equation}
        %         \begin{aligned}
        %           \textit{mIoU}
        %            & =
        %           \frac{1}{C}\sum_{c=1}^{C}\frac{\text{TP}_{c}}{\text{TP}_{c} + \text{FP}_{c} + \text{FN}_{c}}, \\
        %           \textit{Acc}
        %            & =
        %           \frac{\sum_{c=1}^{C}\text{TP}_{c}}
        %           {\sum_{c=1}^{C}(\text{TP}_{c}+\text{FP}_{c})},
        %         \end{aligned}
        %       \end{equation}
        %       where $\text{TP}_{c}$, $\text{FP}_{c}$, and $\text{FN}_{c}$ correspond to the number of true positive, false positive, and false negative predictions for class $c$, and $C$ is the number of classes.
        %       voxel-level intersection over union (IoU). For the semantic scene completion task, we evaluate the IoU of each object classes on both the observed and occluded voxels \cite{song2017semantic}. \bt{TODO by Jianhao}\footnote{\url{https://github.com/PRBonn/semantic-kitti-api/blob/master/evaluate_completion.py}}
        %       \rt{For semantic scene completion, one needs to predict whether a voxel is occupied and its semantic label in the completed scene.}
        % \item \textit{mIOU for completion}: For the scene completion task, we treat all non-empty object class as one category and evaluate IoU of the binary predictions on occluded voxels. Following Firman et al. [3], we do not evaluate on voxels outside the view or the room.
\end{itemize}

\subsubsection{Baseline Methods}
We compare our proposed mapping method with two state-of-the-art TSDF-based mapping methods: \textbf{VoxBlox} and \textbf{VoxField}, which are proposed in \cite{oleynikova2017voxblox} and \cite{pan2022voxfield}, respectively.
Both of them are CPU-based mapping methods, but they do not support semantic mapping and traversability extraction.
Our approach improves the non-projective distance calculation of VoxField by redesigning the weighting strategy.
It also has much difference from VoxField in implementation, including measurement preprocessing, retrieval of visible voxels, and mesh generation using the marching cube algorithm.
The other baselines should be variants of our method that use the original projective distance calculation (\textbf{Ours-Proj}) and does not use the recursive Bayesian Filter in semantic mapping (\textbf{Ours-wo-Bay}), respectively.

\subsubsection{Results on Public Datasets}
\begin{figure}[t]
  \centering
  \subfigure[\textit{SemanticKITTI} $00$ ($0.24km^{2}$)]
  {\label{fig:semantickitti00_mesh}\centering\includegraphics[width=0.22\textwidth]{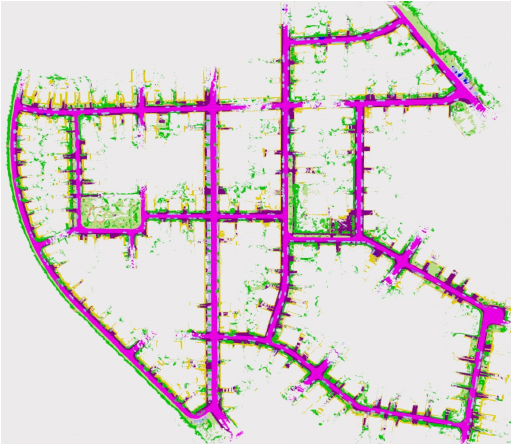}}
  \subfigure[\textit{SemanticKITTI} $05$ ($0.67km^{2}$)]
  {\label{fig:semantickitti05_mesh}\centering\includegraphics[width=0.25\textwidth]{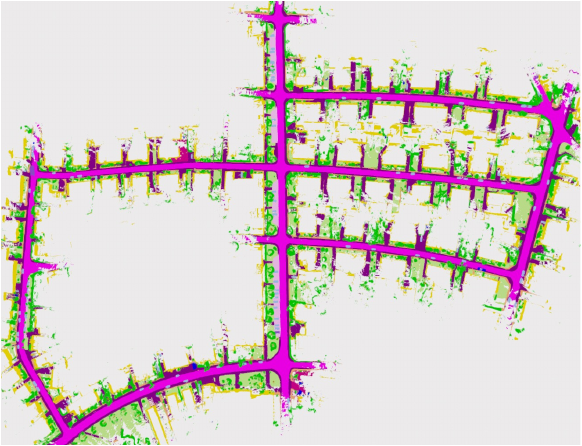}}
  \subfigure[\textit{SemanticUSL} $12$ ($0.014km^{2}$)]
  {\label{fig:semanticusl12_mesh}\centering\includegraphics[width=0.475\textwidth]{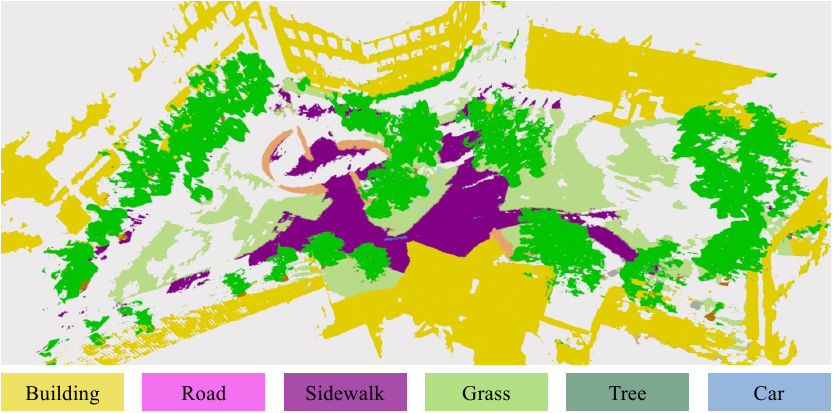}}
  \subfigure[\textit{FusionPortable} \textit{Building\_Day} ($0.019km^{2}$)]
  {\label{fig:20230225building_mesh}\centering\includegraphics[width=0.475\textwidth]{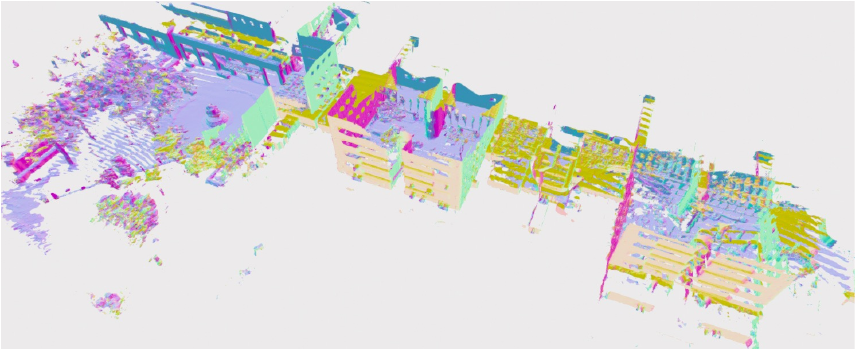}}
  \subfigure[\textit{FusionPortable} \textit{Campus\_Road\_Day} ($0.225km^{2}$)]
  {\label{fig:20230226campusroadday_mesh}\centering\includegraphics[width=0.475\textwidth]{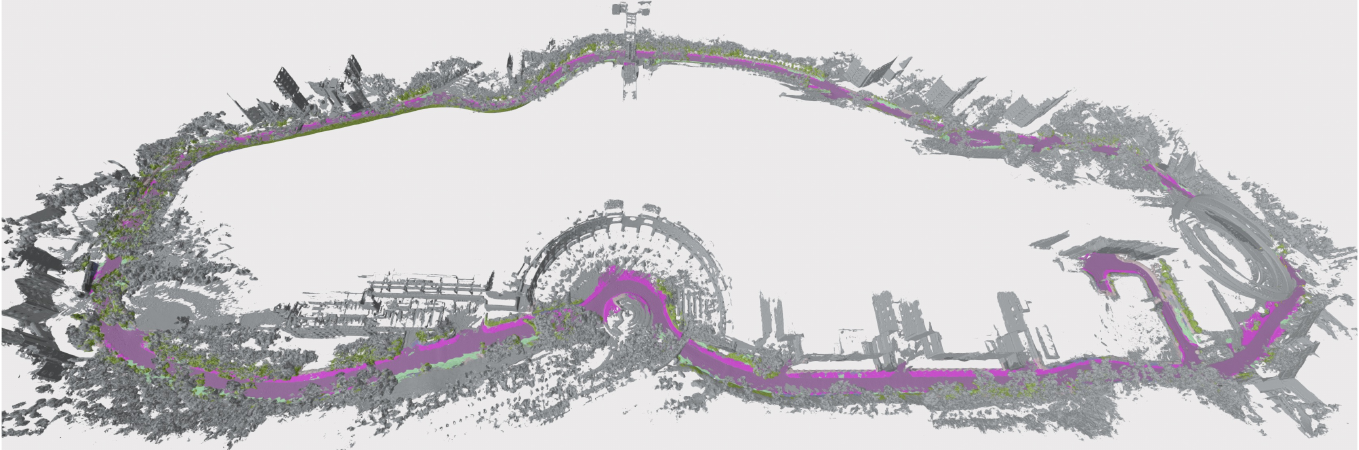}}
  \caption{Results of the global map on four public sequences.
    Semantic labels are shown as colors on maps except for the \textit{FusionPortable} dataset.
    Both (a), (b), and (c) use the same color scheme to (e) does.}
  \label{fig:semantic_mapping_result}
  \vspace{-0.45cm}
\end{figure}

Both \textit{SemanticKITTI} and \textit{SemanticUSL} are two datasets that provide dense annotations for each LiDAR scan.
Sequences $00$\textendash$10$ from \textit{SemanticKITTI} and sequences $03,12,21,32$ from \textit{SemanticUSL} are taken for evaluation since ground-truth labels and maps are provided.
We utilize pretrained Cylinder3D \cite{zhou2020cylinder3d} that is a state-of-the-art LiDAR-only semantic segmentation approach to generate semantic measurements in experiments.
For experiments on \textit{FusionPortable}, we only compute RE, CD, and RC socres since this dataset does not provide semantic annotations.
Sequences \textit{Garden\_Night} (GN), \textit{Canteen\_Night} (CN), \textit{Garden\_Day} (GD), \textit{Canteen\_Day} (CD), \textit{Escalator\_Day} (ED), \textit{Building\_Day} (BD), and \textit{Campus\_Road\_Day} (CRD) are taken in tests.
Fig. \ref{fig:semantic_mapping_result} visualizes the resulting mesh map of several sequences.
Since CRD does not provide the ground-truth map, only qualitative results are shown.

% \input{chapter/exp_table_mapping_result_format_1}
%%%%%%%%%%%%%%%%%%%%%%%%%%%%%%%%%%%%%%%%%%%%%%%%%%%%%%%%%%%%%%%%%%
%%%%%%%%%%%%%%%%%%%%%%%%%%%%%%%%%%%%%%%%%%%%%%%%%%%%%%%%%%%%%%%%%%

Quantitative results on all sequences are given in Table \ref{tab:mapping_result_public_dataset}.
The average computation time reported in the table consists of processing time of these modules: normal image estimation, metric mapping, and semantic mapping.
The non-projective distance calculation is validated to be effective since it improves the construction accuracy of VoxField and ours in terms of RE and CD, as compared with VoxBlox and Ours-Proj.
Due to the advanced implementation of our methods, scores of RE, CD, and time are the highest for the most sequences.
Ours has the lower coverage on \textit{SemanticKITTI} and \textit{SemanticUSL} datasets than VoxBlox does since our method removes unreliable LiDAR points that do not have normal or stay at a large incline angle (especially for ground points), making some voxels empty.
These empty voxels do not have valid distance values, and thus cannot generate mesh.
Data were collected in indoor buildings of the \textit{FusionPortable} dataset
Most of LiDAR points are kept, and thus the RC of ours is large.
Furthermore, both scores of mIoU and Acc of ours are larger than Ours-wo-Bay's, indicating the utility of recursive Bayesian update in maintaining the consistency of semantic information.

We conduct supplementary experiments to assess the influence of factors such as measurement noise, varying view angles, and limited observations on map construction. Due to the page limit, we show these results in the website.

% \input{chapter/exp_table_mapping_result_format_2}
% Please add the following required packages to your document preamble:
% \usepackage{multirow}
% \usepackage{graphicx}

% FusionPortable
% 00 Carteen
% 01 Garden00
% 02 Garden01
% 03 Escalator
% 04 Building

\begin{table*}[t]
  \centering
  \caption{Metric-semantic mapping results in terms of Reconstruction Error ($\downarrow$), Chamfer Distance ($\downarrow$), Reconstruction Coverage ($\uparrow$), Semantic Mapping Score ($\uparrow$), and Computation Time ($\downarrow$).}
  \renewcommand\arraystretch{1.2}
  \renewcommand\tabcolsep{1.0pt}
  \scriptsize
  % \footnotesize
  % \tiny
  \begin{tabular}{|c|c|c|c|c|c|c|c|c|c|c|c|c|c|c|c|c|c|c|c|c|c|c|}
    \Xhline{0.03cm}
    \multirow{2}{*}{Metrics}                     &
    \multirow{2}{*}{Methods}                     &
    \multicolumn{11}{c|}{\textit{SemanticKITTI}} &
    \multicolumn{4}{c|}{\textit{SemanticUSL}}    &
    \multicolumn{6}{c|}{\textit{FusionPortable}}                                                                                                                                                                                                                                                                                               \\ \cline{3-23}
                                                 &
                                                 &
    \rotatebox{0}{00}                            &
    \rotatebox{0}{01}                            &
    \rotatebox{0}{02}                            &
    \rotatebox{0}{03}                            &
    \rotatebox{0}{04}                            &
    \rotatebox{0}{05}                            &
    \rotatebox{0}{06}                            &
    \rotatebox{0}{07}                            &
    \rotatebox{0}{08}                            &
    \rotatebox{0}{09}                            &
    \rotatebox{0}{10}                            &
    \rotatebox{0}{03}                            &
    \rotatebox{0}{12}                            &
    \rotatebox{0}{21}                            &
    \rotatebox{0}{32}                            &
    \rotatebox{0}{GN}                            &
    \rotatebox{0}{CN}                            &
    \rotatebox{0}{GD}                            &
    \rotatebox{0}{CD}                            &
    \rotatebox{0}{ED}                            & \rotatebox{0}{BD}                                                                                                                                                                                                                                                                           \\ \Xhline{0.03cm}
    \multirow{4}{*}{RE$[cm]$}                    &
    VoxBlox                                      &
    $9.7$                                        & $9.4$             & $8.8$       & $8.9$       & $10.4$      & $9.1$       & $9.1$       & $10.0$      & $9.5$       & $8.8$       & $8.9$       & $10.0$      & $9.8$       & $10.1$      & $10.0$      & $9.8$       & $14.8$      & $10.5$      & $14.0$      & $16.0$      & $15.8$
    \\
                                                 &
    VoxField                                     &
    $\bm{7.2}$                                   & $8.0$             & $\bm{6.1}$  & $6.9$       & $7.7$       & $7.6$       & $7.6$       & $7.8$       & $\bm{7.5}$  & $6.8$       & $7.0$       & $8.4$       & $8.1$       & $8.7$       & $8.5$       & $9.9$       & $14.1$      & $10.3$      & $13.5$      & $\bm{15.1}$ & $14.8$
    \\
                                                 &
    Ours-Proj                                    &
    $7.8$                                        & $\bm{7.2}$        & $6.8$       & $6.3$       & $7.0$       & $6.9$       & $7.0$       & $6.8$       & $8.4$       & $6.2$       & $6.1$       & $7.4$       & $7.6$       & $7.3$       & $7.3$       & $9.0$       & $15.1$      & $9.5$       & $13.4$      & $15.5$      & $14.3$
    \\
                                                 &
    Ours                                         &
    $7.5$                                        & $\bm{7.2}$        & $6.4$       & $\bm{5.9}$  & $\bm{6.8}$  & $\bm{6.6}$  & $\bm{6.7}$  & $\bm{6.5}$  & $8.1$       & $\bm{5.8}$  & $\bm{5.8}$  & $\bm{7.1}$  & $\bm{7.1}$  & $\bm{6.7}$  & $\bm{6.9}$  & $\bm{8.8}$  & $\bm{13.5}$ & $\bm{8.9}$  & $\bm{12.6}$ & $\bm{15.1}$ & $\bm{13.8}$
    \\ \Xhline{0.01cm}

    \multirow{4}{*}{CD$[cm]$}                    &
    VoxBlox                                      &
    $6.1$                                        & $7.9$             & $6.9$       & $5.4$       & $7.2$       & $6.5$       & $5.5$       & $5.8$       & $8.8$       & $5.9$       & $5.2$       & $6.1$       & $8.3$       & $9.8$       & $\bm{10.2}$ & $11.2$      & $15.7$      & $10.5$      & $13.9$      & $12.5$      & $12.1$
    \\
                                                 &
    VoxField                                     &
    $6.4$                                        & $\bm{7.7}$        & $7.3$       & $5.3$       & $\bm{6.9}$  & $6.2$       & $4.9$       & $5.4$       & $7.7$       & $5.2$       & $4.8$       & $\bm{5.8}$  & $7.8$       & $9.4$       & $\bm{10.2}$ & $11.5$      & $15.0$      & $12.1$      & $13.8$      & $11.8$      & $12.0$
    \\
                                                 &
    Ours-Proj                                    &
    $6.3$                                        & $8.2$             & $7.6$       & $5.1$       & $7.5$       & $6.0$       & $4.9$       & $5.4$       & $8.0$       & $5.1$       & $4.8$       & $6.0$       & $7.9$       & $9.6$       & $10.5$      & $9.1$       & $16.0$      & $9.9$       & $13.2$      & $12.2$      & $11.2$
    \\
                                                 &
    Ours                                         &
    $\bm{5.9}$                                   & $8.0$             & $\bm{6.5}$  & $\bm{5.0}$  & $7.3$       & $\bm{5.7}$  & $\bm{4.5}$  & $\bm{5.1}$  & $\bm{7.5}$  & $\bm{4.7}$  & $\bm{4.5}$  & $5.9$       & $\bm{7.7}$  & $\bm{9.3}$  & $10.4$      & $\bm{8.2}$  & $\bm{14.0}$ & $\bm{9.2}$  & $\bm{12.7}$ & $\bm{11.1}$ & $\bm{10.6}$
    \\ \Xhline{0.01cm}

    \multirow{4}{*}{RC$[\%]$}                    &
    VoxBlox                                      &
    $\bm{95.3}$                                  & $\bm{88.7}$       & $\bm{94.2}$ & $\bm{96.6}$ & $\bm{92.9}$ & $\bm{91.6}$ & $95.2$      & $\bm{96.6}$ & $86.1$      & $95.3$      & $\bm{97.1}$ & $\bm{94.9}$ & $\bm{87.1}$ & $82.9$      & $\bm{82.1}$ & $70.8$      & $58.2$      & $68.0$      & $63.6$      & $75.2$      & $81.9$
    \\
                                                 &
    VoxField                                     &
    $91.4$                                       & $87.3$            & $89.6$      & $95.3$      & $90.2$      & $91.2$      & $95.6$      & $95.5$      & $87.9$      & $\bm{95.7}$ & $96.5$      & $94.4$      & $87.0$      & $\bm{83.1}$ & $80.5$      & $69.6$      & $59.0$      & $67.7$      & $63.3$      & $75.4$      & $80.0$
    \\
                                                 &
    Ours-Proj                                    &
    $93.2$                                       & $83.5$            & $90.8$      & $94.8$      & $86.7$      & $91.0$      & $95.0$      & $94.1$      & $88.6$      & $94.8$      & $95.1$      & $92.0$      & $85.8$      & $80.1$      & $77.5$      & $78.3$      & $59.4$      & $75.6$      & $64.7$      & $75.4$      & $83.2$
    \\
                                                 &
    Ours                                         &
    $94.0$                                       & $83.7$            & $93.3$      & $94.7$      & $87.0$      & $91.3$      & $\bm{95.9}$ & $94.3$      & $\bm{89.6}$ & $95.4$      & $95.4$      & $92.1$      & $85.9$      & $80.9$      & $77.3$      & $\bm{80.7}$ & $\bm{61.4}$ & $\bm{77.1}$ & $\bm{65.2}$ & $\bm{76.7}$ & $\bm{83.8}$
    \\ \Xhline{0.01cm}

    \multirow{2}{*}{mIoU$[\%]$}                  &
    Ours-wo-Bay                                  &
    $65.7$                                       & $39.0$            & $61.9$      & $63.0$      & $61.2$      & $66.3$      & $62.9$      & $68.3$      & $53.4$      & $64.4$      & $63.3$      & $61.7$      & $43.2$      & $55.9$      & $26.3$      &
    \textendash                                  &
    \textendash                                  &
    \textendash                                  &
    \textendash                                  &
    \textendash                                  &
    \textendash                                                                                                                                                                                                                                                                                                                                \\
                                                 &
    Ours                                         &
    $\bm{76.0}$                                  & $\bm{41.0}$       & $\bm{75.4}$ & $\bm{71.9}$ & $\bm{68.8}$ & $\bm{74.9}$ & $\bm{72.7}$ & $\bm{74.6}$ & $\bm{62.8}$ & $\bm{76.2}$ & $\bm{76.8}$ & $\bm{66.2}$ & $\bm{45.6}$ & $\bm{62.3}$ & $\bm{30.8}$ &
    \textendash                                  &
    \textendash                                  &
    \textendash                                  &
    \textendash                                  &
    \textendash                                  &
    \textendash                                                                                                                                                                                                                                                                                                                                \\ \Xhline{0.01cm}
    \multirow{2}{*}{Acc$[\%]$}                   &
    Ours-wo-Bay                                  &
    $88.7$                                       & $87.5$            & $86.7$      & $88.5$      & $89.4$      & $85.8$      & $85.8$      & $89.0$      & $83.1$      & $86.8$      & $84.2$      & $94.6$      & $93.5$      & $90.2$      & $68.0$      &
    \textendash                                  &
    \textendash                                  &
    \textendash                                  &
    \textendash                                  &
    \textendash                                  &
    \textendash                                                                                                                                                                                                                                                                                                                                \\
                                                 &
    Ours                                         &
    $\bm{92.9}$                                  & $\bm{90.5}$       & $\bm{92.1}$ & $\bm{93.0}$ & $\bm{92.3}$ & $\bm{90.0}$ & $\bm{90.7}$ & $\bm{92.2}$ & $\bm{87.4}$ & $\bm{91.2}$ & $\bm{89.4}$ & $\bm{96.2}$ & $\bm{95.6}$ & $\bm{94.0}$ & $\bm{74.2}$ &
    \textendash                                  &
    \textendash                                  &
    \textendash                                  &
    \textendash                                  &
    \textendash                                  &
    \textendash                                                                                                                                                                                                                                                                                                                                \\
    \Xhline{0.01cm}

    \multirow{4}{*}{Time$[ms]$}                  &
    VoxBlox                                      &
    $200.4$                                      &
    $288.0$                                      &
    $218.5$                                      &
    $242.9$                                      &
    $270.6$                                      &
    $208.8$                                      &
    $286.9$                                      &
    $181.9$                                      &
    $231.8$                                      &
    $235.5$                                      &
    $185.5$                                      &
    $324.2$                                      &
    $368.8$                                      &
    $293.6$                                      &
    $267.6$                                      &
    $327.7$                                      &
    $282.3$                                      &
    $331.3$                                      &
    $288.2$                                      &
    $158.4$                                      &
    $271.4$                                                                                                                                                                                                                                                                                                                                    \\
                                                 &
    VoxField                                     &
    $130.5$                                      &
    $157.0$                                      &
    $141.2$                                      &
    $188.0$                                      &
    $154.2$                                      &
    $131.9$                                      &
    $157.2$                                      &
    $115.9$                                      &
    $144.4$                                      &
    $140.8$                                      &
    $115.1$                                      &
    $188.0$                                      &
    $214.6$                                      &
    $164.7$                                      &
    $157.7$                                      &
    $329.6$                                      &
    $281.4$                                      &
    $318.0$                                      &
    $286.5$                                      &
    $162.3$                                      &
    $281.9$                                                                                                                                                                                                                                                                                                                                    \\
                                                 &
    Ours ($3080$Ti)                              &
    $\bm{2.4}$                                   &
    $\bm{6.5}$                                   &
    $\bm{2.5}$                                   &
    $\bm{3.7}$                                   &
    $\bm{4.3}$                                   &
    $\bm{3.1}$                                   &
    $\bm{4.1}$                                   &
    $\bm{2.9}$                                   &
    $\bm{2.8}$                                   &
    $\bm{3.6}$                                   &
    $\bm{2.8}$                                   &
    $\bm{5.4}$                                   &
    $\bm{6.8}$                                   &
    $\bm{6.3}$                                   &
    $\bm{6.0}$                                   &
    $\bm{1.6}$                                   &
    $\bm{1.6}$                                   &
    $\bm{1.6}$                                   &
    $\bm{1.5}$                                   &
    $\bm{1.4}$                                   &
    $\bm{1.7}$
    \\

                                                 &
    Ours (ORIN)                                  &
    $17.2$                                       &
    $31.9$                                       &
    $17.9$                                       &
    $24.2$                                       &
    $26.0$                                       &
    $23.0$                                       &
    $25.6$                                       &
    $21.0$                                       &
    $20.3$                                       &
    $24.7$                                       &
    $19.9$                                       &
    $29.0$                                       &
    $34.4$                                       &
    $24.9$                                       &
    $22.6$                                       &
    $16.7$                                       &
    $16.2$                                       &
    $16.2$                                       &
    $16.4$                                       &
    $11.5$                                       &
    $16.8$
    \\
    \Xhline{0.03cm}
  \end{tabular}
  \label{tab:mapping_result_public_dataset}
  \vspace{-0.3cm}
\end{table*}

\begin{table}[t]
  \centering
  \caption{Computation Time $[ms]$ on the \textit{SemanticKITTI} $00$ and acceleration ratio compared with VoxField.}
  \renewcommand\arraystretch{1.2}
  \renewcommand\tabcolsep{3pt}
  \scriptsize
  \begin{tabular}{r|rrr|r}
    \Xhline{0.03cm}
    Methods                   & Normal Image & Metric Map.     & Semantic Map. & Mesh Generation \\
    \Xhline{0.03cm}
    VoxBlox                   & \textendash  & $200.4\pm 30.6$ & \textendash   & $111.2\pm 41.1$ \\ \hline
    VoxField                  & $6.3\pm 2.0$ & $124.2\pm 17.6$ & \textendash   & $99.7\pm 25.0$  \\ \hline
    Ours ($3080$Ti)           &
    \begin{tabular}[r]{@{}r@{}}$\bm{0.2}\pm 0.1$\\$(\times 31.4)$\end{tabular} &
    \begin{tabular}[r]{@{}r@{}}$\bm{1.0}\pm 0.2$\\$(\times 124.2)$\end{tabular} &
    $\bm{1.0}\pm 0.2$         &
    \begin{tabular}[r]{@{}r@{}}$\bm{32.3}\pm 7.7$\\$(\times 3.1)$\end{tabular}
    \\ \hline
    Ours (ORIN)               &
    \begin{tabular}[r]{@{}r@{}}$0.7\pm 0.3$\\$(\times 9.0)$\end{tabular} &
    \begin{tabular}[r]{@{}r@{}}$9.3\pm 0.7$\\$(\times 13.4)$\end{tabular} &
    $8.0\pm 0.9$              &
    \begin{tabular}[r]{@{}r@{}}$232.8\pm 83.0$\\$(\times 0.4)$\end{tabular}
    \\
    \Xhline{0.03cm}
  \end{tabular}
  \label{tab:computation_time}
  \vspace{-0.4cm}
\end{table}

\begin{comment}
\begin{table}[t]
  \centering
  \caption{Computation Time $[ms]$ on the \textit{SemanticKITTI} $00$ and acceleration ratio compared with VoxField.}
  \renewcommand\arraystretch{1.2}
  \renewcommand\tabcolsep{3pt}
  \scriptsize
  \begin{tabular}{r|rrrr}
    \Xhline{0.03cm}
    Steps         & VoxBlox     & VoxField    & Ours                      & Ours (ORIN)               \\
    \Xhline{0.03cm}
    Normal Image  & \textendash & $6\pm 2$    & $\bm{0.2}\pm 0.1\ (31.4)$ & $\bm{0.6}\pm 0.1\ (31.4)$ \\
    Metric Map.   & $200\pm 31$ & $124\pm 18$ & $\bm{1}\pm 0.2\ (124.2)$  & $\bm{10}\pm 0.2\ (124.2)$ \\
    Semantic Map. & \textendash & \textendash & $1\pm 0.2$                & $12\pm 2$                 \\
    \Xhline{0.01cm}
    Mesh Gene.    & $111\pm 41$ & $100\pm 25$ & $\bm{32}\pm 8\ (3.1)$     & $\bm{32.3}\pm 7.7\ (3.1)$ \\
    \Xhline{0.03cm}
  \end{tabular}
  \label{tab:computation_time}
  \vspace{-0.4cm}
\end{table}
\end{comment}

\subsubsection{Qualitative Results on Self-Collected Datasets}
We use the mapping device to collect data in the campus.
We collected two typical sequences that contain objects including roads, sidewalks, terrain, vegetations, vehicles, and buildings, which are appropriate to test our metric-semantic mapping method.
Fig. \ref{fig:semantic-map} visualizes the resulting metric-semantic map that is aligned with the statllite image.
Colors of each point of the map indicate the label.
Due to the limited field of view of the camera, plenty of points are not annotated.

\subsubsection{Timing}
Table \ref{tab:computation_time} reports the detailed computation time regarding each step, with comparison of VoxBlox and VoxField.
We take the typical sequence $00$ of \textit{SemanticKITTI} as an example which has over $0.24 km^{2}$.
Most of computations of mapping are done in GPUs and very fast, even on the Jetson ORIN.
Computing normals on a range image requires around $0.2ms$.
The metric mapping module that processes each new frame takes an average of $1.0ms$, including gathering visible voxels by ray tracing as well as updating their distance and weight.
The semantic mapping module needs to find visible voxels and update their class probabilities via. the Bayesian filter, costing around $1.0ms$.
Our method with the $3080$Ti GPU only takes an average of $32.3ms$ to update the global metric-semantic mesh at a fixed frequency.

% OLjjh1994
%%%%%%%%%%%%%%%%%%%%%%%%%%%%%%%%%%%%%%%%%%%%%%%%%%%%%%%%%%%%%%%%%%
\subsection{Point-Goal Navigation Experiments}
\label{sec:experiment_navigation}

\begin{figure}[t]
    \centering
    \subfigure[Semantic map (seq.$00$)]
    {\label{fig:semantics-sequence00}\centering\includegraphics[width=0.21\textwidth]{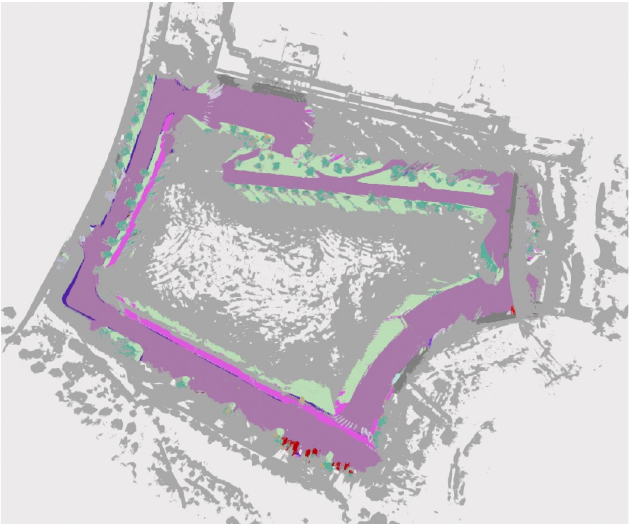}}
    \subfigure[Aligned on a top-view image]
    {\label{fig:topview-sequence00}\centering\includegraphics[width=0.24\textwidth]{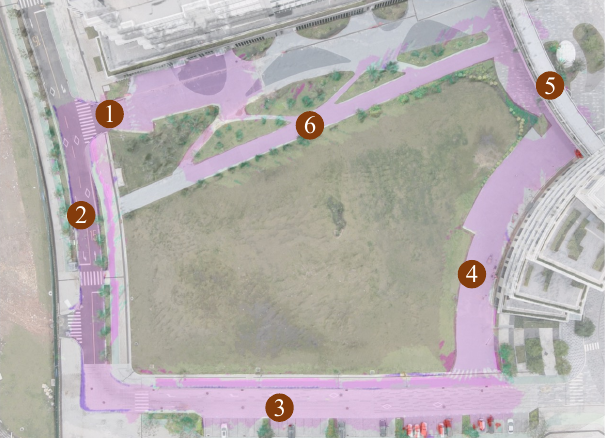}}
    \subfigure[Semantic map (seq.$01$)]
    {\label{fig:semantics-sequence01}\centering\includegraphics[width=0.2225\textwidth]{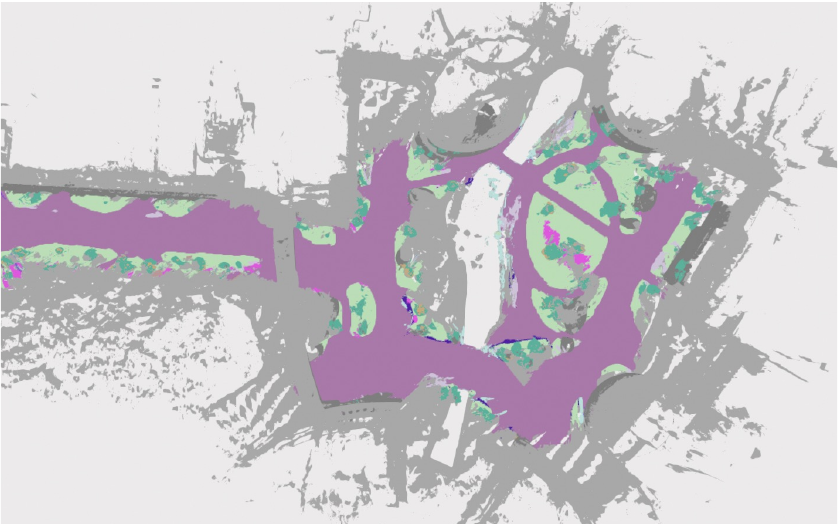}}
    \subfigure[Aligned on a top-view image]
    {\label{fig:topview-sequence01}\centering\includegraphics[width=0.22725\textwidth]{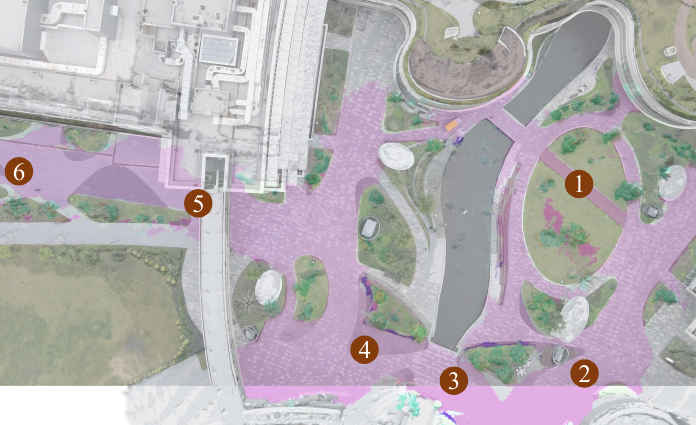}}
    \caption{Semantic maps are created from the self-collected datasets: sequence $00$ (top) and sequence $01$ (bottom). Maps are manually aligned with images to show the specific meaning of labels. Traversable regions are then extracted from these maps. In navigation experiments, we command a vehicle to drive through goals that are marked in (b) and (d). Third-view pictures that show how the vehicle drives are presented in Fig. \ref{fig:navigate-test}.}
    \label{fig:semantic-map}
    \vspace{-0.3cm}
\end{figure}

%%%%%%%%%%%%%%%%%%%%%%%%%%%%%%%%%%%%%%%%%%%%%%%%%%%%%%%%%%%%%%%%%%%%%%%%%
\begin{figure*}[t]
  \centering
  \subfigure[]
  {\label{fig:mesh-terrain-heightdiff-sequence00}\centering\includegraphics[width=0.195\textwidth]{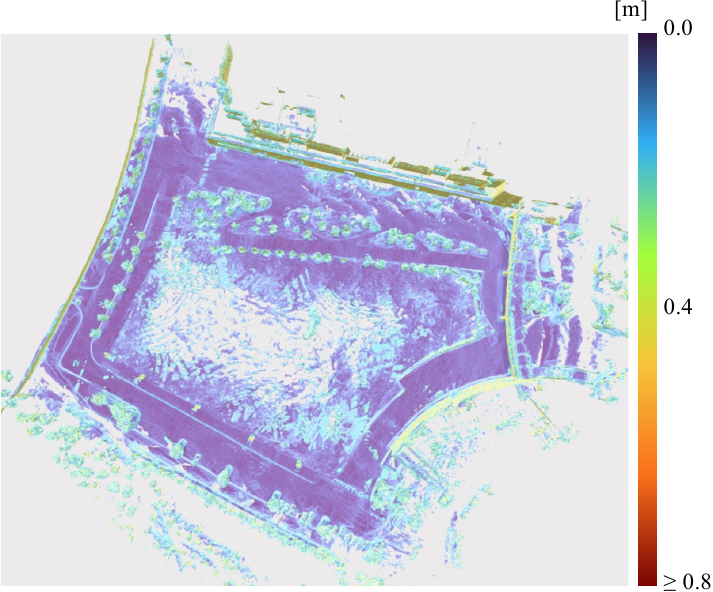}}
  \subfigure[]
  {\label{fig:mesh-terrain-steepness-sequence00}\centering\includegraphics[width=0.195\textwidth]{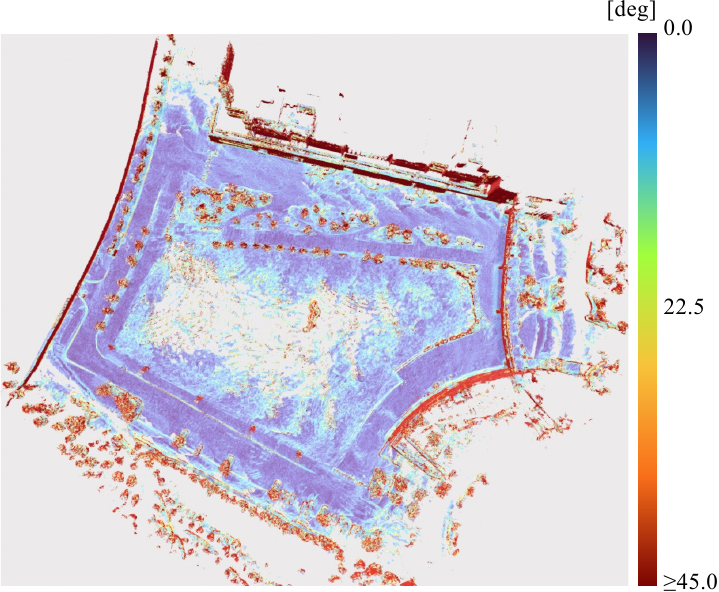}}
  \subfigure[]
  {\label{fig:mesh-terrain-roughness-sequence00}\centering\includegraphics[width=0.195\textwidth]{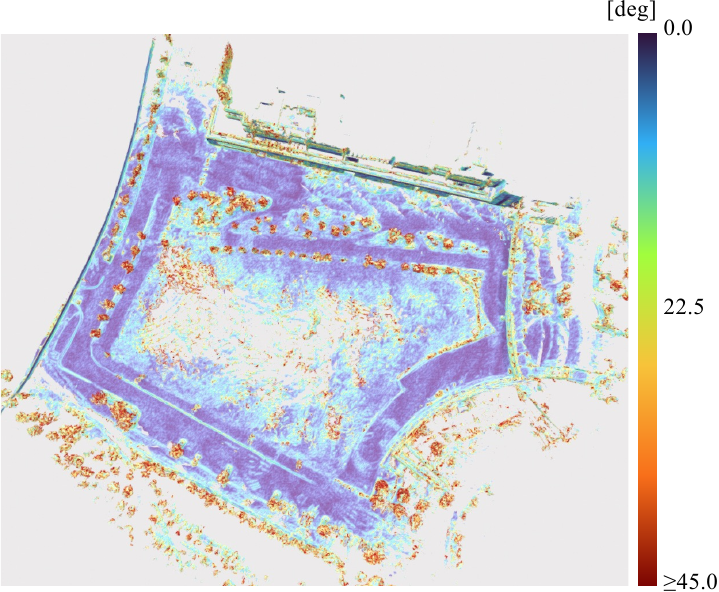}}
  \subfigure[]
  {\label{fig:mesh-terrain-occupancy-wsemantics-sequence00}\centering\includegraphics[width=0.174\textwidth]{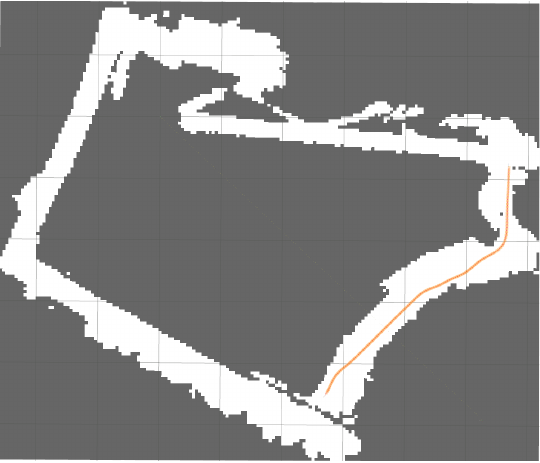}}
  \subfigure[]
  {\label{fig:mesh-terrain-occupancy-wosemantics-sequence00}\centering\includegraphics[width=0.166\textwidth]{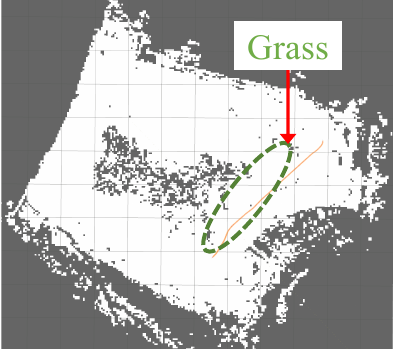}}
  \subfigure[]
  {\label{fig:mesh-terrain-heightdiff-sequence01}\centering\includegraphics[width=0.19\textwidth]{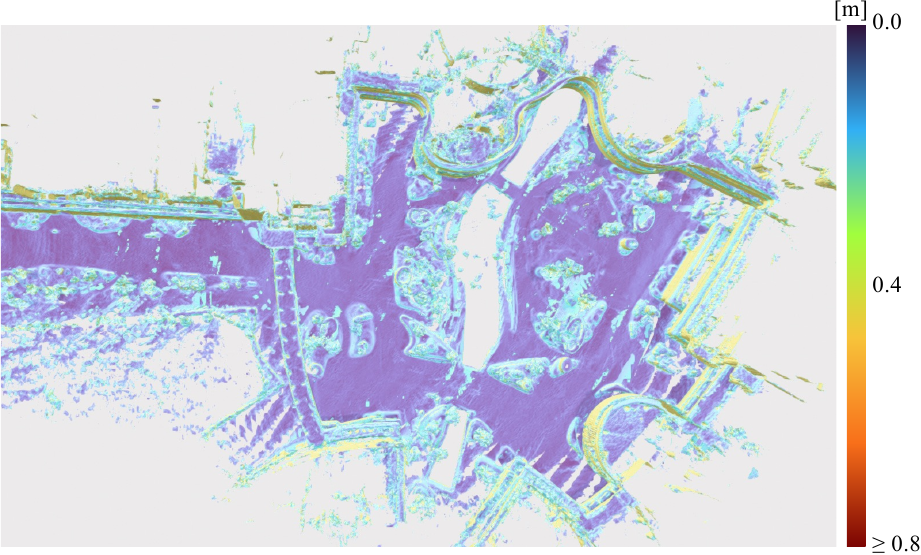}}
  \subfigure[]
  {\label{fig:mesh-terrain-steepness-sequence01}\centering\includegraphics[width=0.19\textwidth]{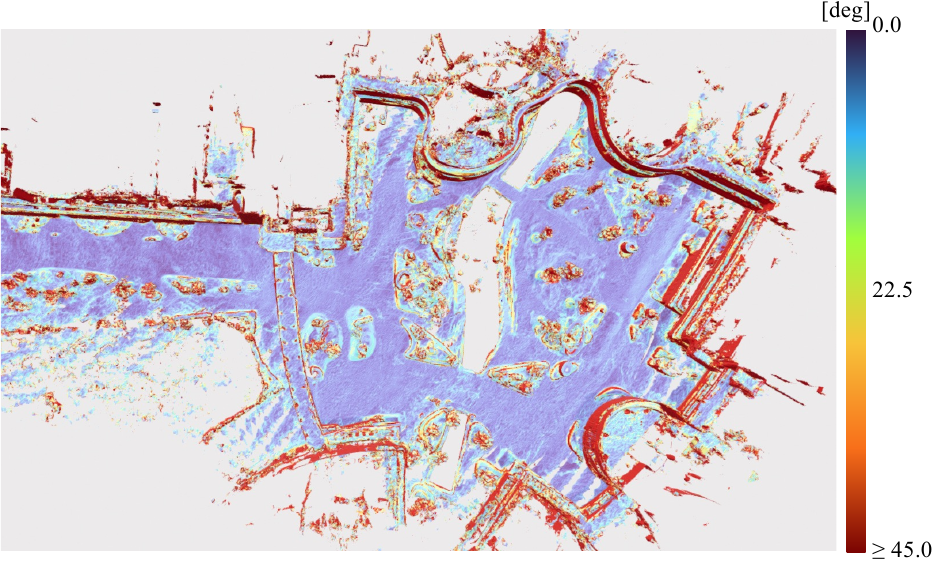}}
  \subfigure[]
  {\label{fig:mesh-terrain-roughness-sequence01}\centering\includegraphics[width=0.19\textwidth]{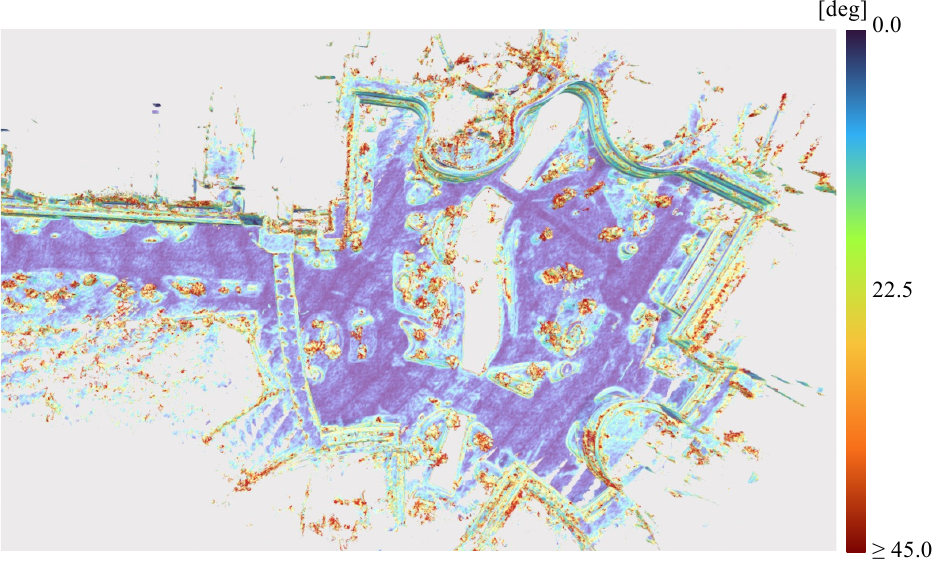}}
  \subfigure[]
  {\label{fig:mesh-terrain-occupancy-wsemantics-sequence01}\centering\includegraphics[width=0.19\textwidth]{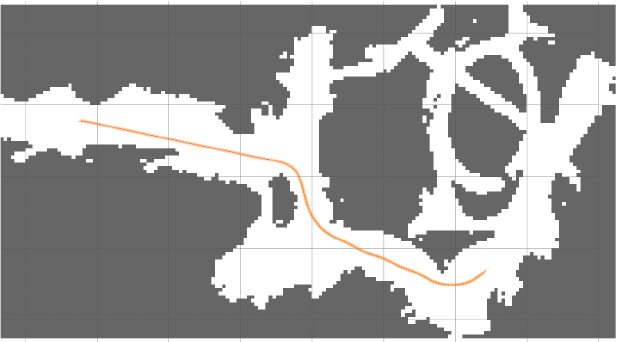}}
  \subfigure[]
  {\label{fig:mesh-terrain-occupancy-wosemantics-sequence01}\centering\includegraphics[width=0.16\textwidth]{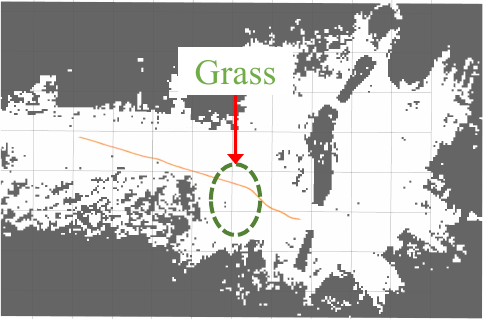}}
  \caption{Visualization of geometric properties of the resulting metric-semantic mesh map and projected 2D occupancy maps for navigation on sequence$00$ (top) and sequence $01$ (bottom): (a)(f) height difference, (b)(g) steepness, (c)(h) roughness, (d)(i) occupancy map using semantic information, and (e)(j) occupancy map without using semantic information. The yellow lines in (d)(i) and (e)(j) indicate the found navigation paths, where the path in (d)(i) does not intersect with untraversable regions.}
  \label{fig:mesh-terrain-sequence}
\end{figure*}
%%%%%%%%%%%%%%%%%%%%%%%%%%%%%%%%%%%%%%%%%%%%%%%%%%%%%%%%%%%%%%%%%%%%%%%%%

%%%%%%%%%%%%%%%%%%%%%%%%%%%%%%%%%%%%%%%%%%%%%%%%%%%%%%%%%%%%%%%%%%%%%%%%%
\begin{figure*}[t]
    \centering
    \subfigure[Navigation in the region that is covered by the sequence $00$.]
    {\label{fig:navigate-test-sequence00}\centering\includegraphics[width=0.91\textwidth]{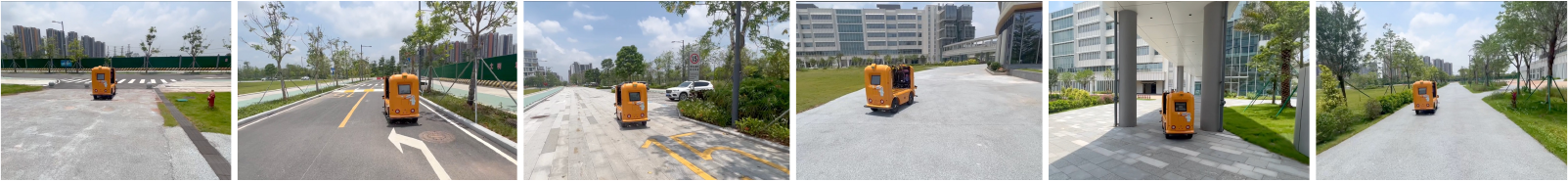}}
    \subfigure[Navigation in the region that is covered by the sequence $01$.]
    {\label{fig:navigate-test-sequence01}\centering\includegraphics[width=0.905\textwidth]{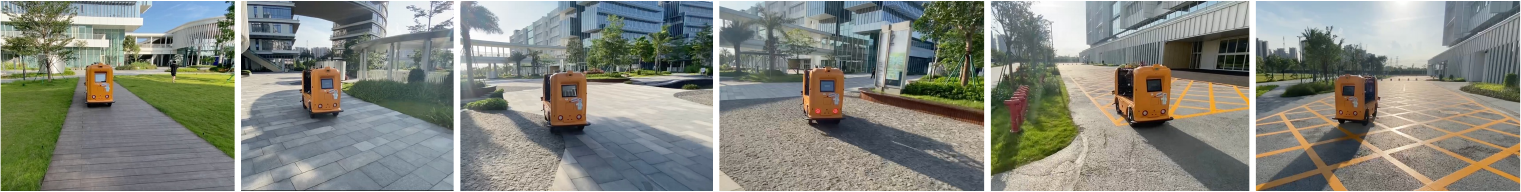}}
    \caption{Without driving into grassland and sidewalks, the vehicle successfully navigate via. regions that are covered by the self-collected dataset after being given a set of goal points. These pictures are capture in places which are indicated on maps shown in Fig. \ref{fig:semantic-map}.}
    \label{fig:navigate-test}
    \vspace{-0.4cm}
\end{figure*}
%%%%%%%%%%%%%%%%%%%%%%%%%%%%%%%%%%%%%%%%%%%%%%%%%%%%%%%%%%%%%%%%%%%%%%%%%

%%%%%%%%%%%%%%%%%%%%%%%%%%%%%%%%%%%%%%%%%%%%%%%%%%%%%%%%%%%%%%%%%%%%%%%%%
% \begin{figure*}[t]
%   \centering
%   \includegraphics[width=0.99\textwidth]{figure/experiment/navigation_test_sequence01-crop.pdf}
%   \caption{Autonomous navigation in the scenario whose metric-semantic map is built on the sequence$01$.}
%   \label{fig:navigate-test-sequence01}
% \end{figure*}
%%%%%%%%%%%%%%%%%%%%%%%%%%%%%%%%%%%%%%%%%%%%%%%%%%%%%%%%%%%%%%%%%%%%%%%%%

\subsubsection{Results of Traversability Extraction}
After computing geometric properties (\textit{i.e.,} ``height difference'', ``steepness'', and ``roughness'') of the resulting metric-semantic mesh map for the two self-collected sequences, we visualize these values in Fig. \ref{fig:mesh-terrain-sequence}.
Considering the vehicle's mobility, objects that are not traversable such as cars, buildings, and trees are easily distinguished and filtered out by setting thresholds.
But for the sidewalk (designed for pedestrians) and grassland which are not traversable for vehicles have to be distinguished by semantic information.
By combining all geometric and semantic information, we obtain the 2D occupancy map, as shown in Fig. \ref{fig:mesh-terrain-occupancy-wsemantics-sequence00} and Fig. \ref{fig:mesh-terrain-occupancy-wsemantics-sequence01}, respectively.

\subsubsection{Results of Real-World Navigation}
To demonstrate the practical application of our occupancy maps in motion planning and validate their effectiveness, we conducted a preliminary experiment.
We designated start and end points on the map, with the resulting paths visualized in Fig. \ref{fig:mesh-terrain-occupancy-wsemantics-sequence00} and Fig. \ref{fig:mesh-terrain-occupancy-wosemantics-sequence01}. This test reveals a critical insight: paths generated without integrating semantic information inadvertently cross through grassland areas.
Such terrains, characterized by their uneven nature, pose significant navigational hazards, underscoring the vehicle's risk of becoming ensnared.
This observation starkly highlights the necessity of semantic insights to distinguish between traversable and non-traversable regions, thereby ensuring the safety and reliability of the navigation paths chosen.

We extended our research to include practical applications by deploying the map on a real-world vehicle.
Demonstrated in Fig. \ref{fig:topview-sequence00} and Fig. \ref{fig:topview-sequence01}, the vehicle was tasked with completing two navigation tests based on a series of predefined goal points. The motion planner successfully identified collision-free trajectories, enabling the vehicle to navigate the prescribed paths effectively.
Visual evidence of these navigation tests is captured in third-person photographs, as showcased in Fig. \ref{fig:navigate-test}, with the vehicle achieving an average speed of approximately $3m/s$. For a comprehensive view of these tests, we invite readers to view the accompanying demonstration video.

\subsubsection{Discussion}
Our proposed system represents a robust and efficient mapping solution, characterized by its high computational performance and versatile design.
Crafted with modularity at its core and seamlessly integrated with the Robot Operating System (ROS), it offers unparalleled flexibility for customization to meet diverse application needs.
Empirical evidence from our experiments underscores the metric-semantic map's superiority in facilitating enhanced visualization, precise localization, and effective traversability analysis for navigation.
By embedding semantic information that reflects human knowledge, the system adeptly distinguishes between drivable and non-drivable areas, such as sidewalks and grasslands. This feature not only elevates the safety of autonomous navigation but also significantly diminishes the human effort required for robotic system implementation, marking a notable advancement over previous efforts \cite{liu2021role}.

\section{Conclusion}
\label{sec:conclusion}

In this paper, we introduce an online metric-semantic mapping system tailored for autonomous navigation, featuring LiDAR-IMU-Visual odometry, image-based semantic segmentation, TSDF-based mapping, and extraction of traversable areas.
We further integrate this mapping with a navigation system, enhancing map-based localization and motion planning.
Our evaluation includes extensive mapping and point-to-point navigation tests across $24$ sequences from both public and proprietary datasets in a campus setting.

Despite its strengths, our system faces limitations, notably in GPU memory reliance, which challenges city-scale mapping scalability (\textit{e.g.}, AutoMerge \cite{yin2023automerge}). A potential remedy is a submap approach, balancing voxel storage between GPU for immediate access and CPU for less active data. Additionally, the absence of loop correction introduces drift over time, an issue that could be alleviated by integrating submap techniques and mesh deformation optimizations for map correction (\textit{e.g.}, Kimera \cite{tian2022kimera}). Lastly, maintaining semantic features' spatio-temporal consistency poses difficulties, with potential solutions hinted at in kernel-based methods (\textit{e.g.}, \cite{gan2022multitask}).

\section*{Acknowledgement}
The authors thank Qingwen Zhang, Yingbing Chen, Mingkai Tang, Xiangcheng Hu, Hexiang Wei, and Tianshuai Hu for their suggestions on the mapping system and real-world navigation experiments.
They also thank the Robotics and Autonomous System (ROAS) at the Hong Kong University of Science and Technology (GZ) for providing the experimental site, and gratefully acknowledge ChatGPT for polishing the manuscript.

\bibliographystyle{IEEEtran}
%%%%%%%%%%%%%%%%%%%%%%%%%%%%%%%%%%%%%%%%%%%%%%%%%%%%%%%%%%%%%%%%%%%%%%%

%%%%%%%%%%%%%%%%%%%%%%%%%%%%%%%%%%%%%%%%%%%%%%%%%%%%%%%%%%%%%%%%%%%%%%%%%%%%%%%%
\bibliography{reference}
\vspace{-0.9cm}

\end{document}